\documentclass[manuscript,nonacm]{acmart}

\AtBeginDocument{%
  \providecommand\BibTeX{{%
      \normalfont B\kern-0.5em{\scshape i\kern-0.25em b}\kern-0.8em\TeX}}
}

\usepackage{booktabs}
\usepackage{xcolor}
\usepackage{tikz}
\usepackage{pgfplots}
\usepackage{makecell}
\usepackage[inline]{enumitem}
\usepackage{subcaption}
\usepackage{graphicx}
\usepackage{backref}
\usepackage{pifont}
\usepackage{soul} 
\usepackage{hyperref}

\newcommand{\cmark}{\ding{51}}%
\newcommand{\xmark}{\ding{55}}%

\usetikzlibrary{arrows,shapes,positioning,shadows,trees}

\tikzset{
  basic/.style  = {draw, text width=2cm, drop shadow, font=\sffamily, rectangle},
  root/.style   = {basic, rounded corners=2pt, thin, align=center,
                   fill=green!30},
  level 2/.style = {basic, rounded corners=6pt, thin,align=center, fill=green!60,
                   text width=8em},
  level 3/.style = {basic, thin, align=left, fill=pink!60, text width=6.5em}
}

\setcopyright{none}
\copyrightyear{}
\acmDOI{}

\acmJournal{CSUR}
\acmVolume{X}
\acmNumber{Y}
\acmArticle{Z}
\acmMonth{13}

\def\mySectionSymbol~{\S{}}

\begin{document}

\title{Machine Learning at the Network Edge: A Survey}
%
\author{M.G.~Sarwar Murshed}
\email{murshem@clarkson.edu}
\affiliation{%
  \institution{Clarkson University}
  \city{Potsdam}
  \state{NY}
  \postcode{13699}
}

\author{Christopher Murphy}
\affiliation{%
  \institution{SRC, Inc.}
  \city{North Syracuse}
  \state{NY}}
\email{cmurphy@srcinc.com}

\author{Daqing Hou}
\email{dhou@clarkson.edu}
\affiliation{%
  \institution{Clarkson University}
  \city{Potsdam}
  \state{NY}
  \postcode{13699}
}

\author{Nazar Khan}
\email{nazarkhan@pucit.edu.pk}
\affiliation{%
 \institution{Department of Computer Science, University of the Punjab}      
 \city{Lahore}
 \state{Punjab}
 \country{Pakistan}
 }
 
\author{Ganesh Ananthanarayanan}
\email{ga@microsoft.com}
\affiliation{%
  \institution{Microsoft Research}
  \city{Redmond}
  \state{WA}}

\author{Faraz Hussain}
\orcid{0000-0001-8971-1850}
\email{fhussain@clarkson.edu}
\affiliation{%
  \institution{Clarkson University}
  \city{Potsdam}
  \state{NY}
  \postcode{13699}
}

%
\renewcommand{\shortauthors}{Murshed, et al.}

\begin{abstract}
  Resource-constrained IoT devices, such as sensors and actuators,
  have  become ubiquitous in recent years.
  This has led to the generation of large quantities of data in real-time, which is
  an appealing target for AI systems. 
  However, deploying machine learning models on such end-devices is nearly impossible.
  A typical solution involves offloading data to external computing systems (such as cloud servers)
  for further processing
  but this worsens latency, leads to increased communication costs, and adds to privacy concerns.
  To address this issue, efforts have been made to
  place additional computing devices at the edge of the network,
  i.e close to the IoT devices where the  data is generated.  
  Deploying machine learning systems on such edge computing devices alleviates the above issues
  by allowing computations to be performed close to the data sources.
  This survey describes major research efforts where machine learning systems have been
  deployed at the edge of computer networks, focusing on the operational aspects 
  including compression techniques, tools, frameworks, and hardware used in
  successful applications of intelligent edge systems.
\end{abstract}


\maketitle

\section{INTRODUCTION}
Due to the explosive growth of wireless communication technology, the number of Internet of Things (IoT) devices has increased dramatically in recent years. It has been estimated that by 2020, more than 25 billion devices will have been connected to the Internet \cite{MAHDAVINEJAD2018161} and the potential economic impact of the IoT will be \$3.9 trillion to \$11.1 trillion annually by 2025 \cite{Manyika15IoTReport}. IoT devices typically have limited computing power and small memories.
Examples of such resource-constrained IoT devices include sensors, microphones, smart fridges, and smart lights.
IoT devices and sensors continuously generate large amounts of data, which
is of critical importance to many modern technological applications such as autonomous vehicles.

One of the best ways to extract information and make decisions from this data is to feed those data to a machine learning (ML) system.
Unfortunately, limitations in the computational capabilities of resource-constrained devices inhibit
the deployment of ML algorithms on them.
So, the data is offloaded to remote computational infrastructure, most commonly cloud servers, where computations are performed.
Transferring raw data to cloud servers increases communication costs, causes delayed system response,
and makes any private data vulnerable to compromise.
To address these issues, it is natural to consider processing data closer to its sources
and transmitting only the necessary data to remote servers for further processing \cite{6472113}.

\begin{figure*}[ht]
\center
  \fbox{\includegraphics[width=0.3\linewidth]{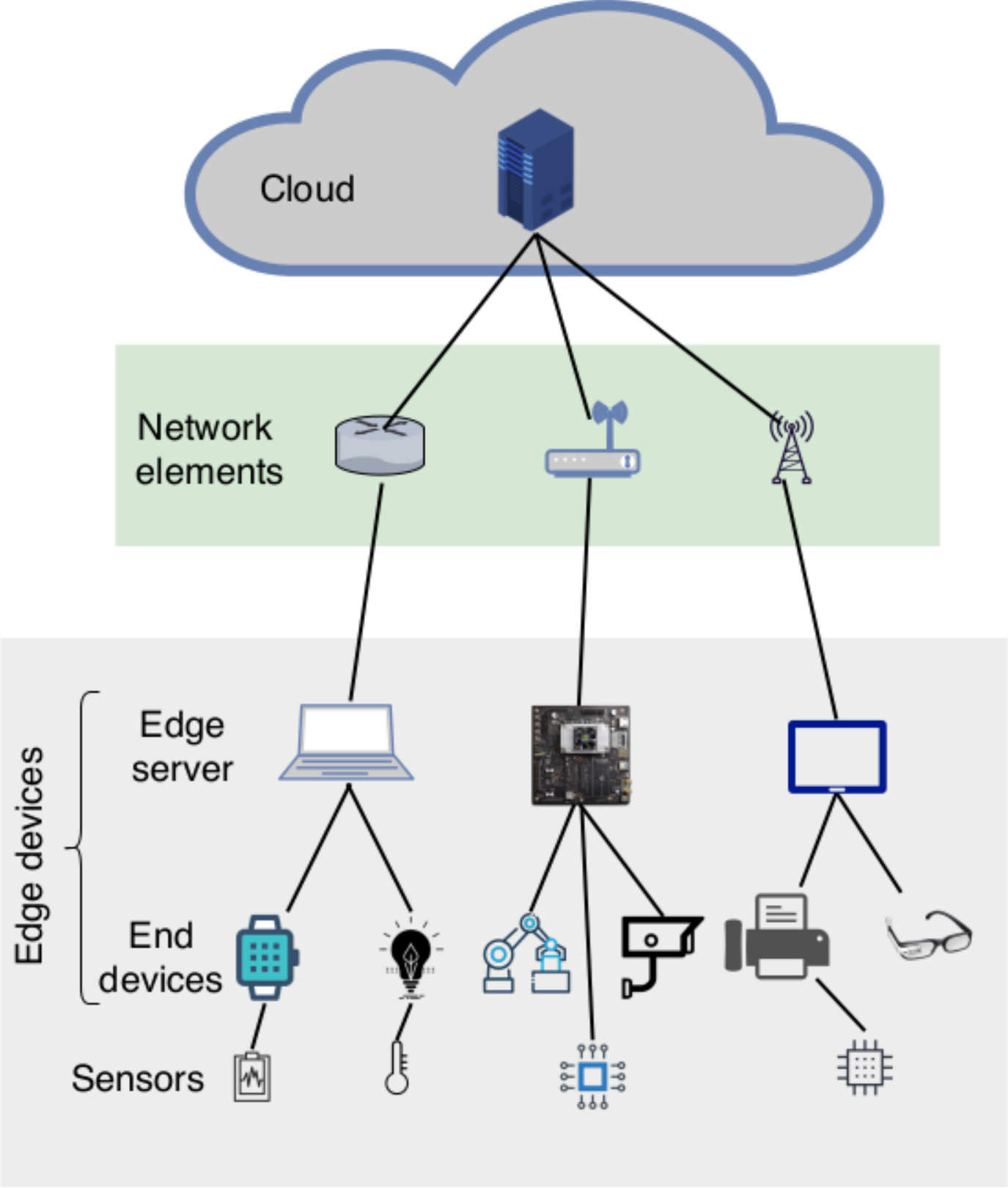}}
  \caption{A general overview of the edge computing architecture: 
    an edge-device is  either an end-device or an
edge-server.   An end-device is one with limited computing capability
    on which data can be produced and some computationally inexpensive (but useful) work can be performed.
    Computationally incapable sensors (last row) just produce data and are connected to either an end-device or an edge-server. 
    The computational capability of an edge-server is also limited,
    but it has more computing resources than an end-device so it can be used for  more complex tasks.
    Network elements are used to connect edge computing architecture to the cloud server.}
  \label{fig:edgeArch}
\end{figure*}

\emph{Edge computing} refers to computations being performed  as close to
data sources as possible, instead of on  far-off, remote locations \cite{7807196, butler2017whatisedge}. 
This is achieved by adding edge computing devices close to the resource-constrained devices where data is generated.
{\autoref{fig:edgeArch} provides an overview of the edge computing architecture.
  In the rest of the paper, the term `edge-device' is used to refer to either an end-device or an edge-server.}

These edge-devices possess both computational and communication capabilities \cite{7469991}. 
For example, an embedded device such as an Nvidia Jetson NANO could serve as an edge-device if it takes data from a camera on a robotic arm and performs data processing tasks that are used to determine the next movement of the arm.
Since edge-devices have limited compute power, energy consumption is a critical factor \cite{magid2019image}
and computations too intense for edge-devices are sent over to more powerful remote servers.

As edge computing has emerged as an important paradigm for IoT based systems \cite{Shi2016EdgeCV},
attempts are being made to use
devices at the network edge to do as much computation as possible,
instead of only using the cloud for processing data \cite{DBLP:journals/corr/abs-1907-06040}.
For example, 
Floyer studied data management and processing costs of a remote wind-farm using a cloud-only system versus a combined edge-cloud system \cite{FloyerEdgeIoT}.  
The wind-farm consisted of several data producing sensors and devices such as video surveillance cameras,
security sensors, access sensors for all employees, and sensors on  wind-turbines.
The edge-cloud system turned out to be $36\%$ less expensive costing only $\$28, 927$ as opposed to
the cloud-only system which cost $\$80, 531$.
Also, the volume of data required to be transferred
was observed to be $96\%$ less, compared to the cloud-only system.
Similar examples include a video analytics system for autonomous drones where an edge computing system is used to save bandwidth \mbox{\cite{Wang2018VideoAnalytics}}, which reduces the need to transfer video data to the cloud by up to 10 times.
  In another study, Murshed et al. deployed a resource-aware deep learning system using edge devices where all image data is processed at the network edge \mbox{\cite{9356355}}.   By processing all data locally, this system not only preserves the privacy of the data but also saves bandwidth.
  Further use cases of edge computing in deep learning systems and their benefits are discussed in a dedicated
  section on applications (\autoref{sec:EdgeMLApplications}).

\noindent Performing computations at the network edge  has several advantages:

\begin{itemize}[topsep=0pt,itemsep=0ex,partopsep=1ex,parsep=1ex]
  \small
\item The volume of data needed to be transferred to a central computing location is reduced because
  some of it is processed by edge-devices.
\item The physical proximity of edge-devices to the data sources makes it possible to achieve lower latency
  which improves real-time data processing performance.
\item For the cases where data must be processed remotely, edge-devices can be used to discard personally identifiable information (PII) prior to data transfer,
  thus enhancing user privacy and security. 
\item Decentralization can make systems more robust by providing transient services during a network failure or cyber attack.
\end{itemize}

Major technology firms, the defense industry, and the open source community
have all been at the forefront in investments in edge technology \cite{johnson2019neural}:

\begin{itemize}[topsep=0pt,itemsep=0ex,partopsep=1ex,parsep=1ex]
    \small
\item Researchers have developed a new architecture called Agile Condor
which uses ML algorithms to perform real-time computer vision tasks (e.g. video, image processing,
and pattern recognition) \cite{8324277}.
With the processing capabilities of 7.5 teraflops, this architecture can run the latest deep learning-based object detection architectures like Faster RCNN, YOLO in real-time \mbox{\cite{8547797}}. Agile Condor can be used for automatic target recognition (ATR) at the network edge, near the data sources. A successful demonstration to identify targets of interest has been shown in 2018, where Agile Condor successfully identified a convoy in the real world.
  However, further analysis is required to validate and verify results against ground truth datasets.

\item Microsoft recently introduced
  HoloLens 2\footnote{\url{https://www.microsoft.com/en-us/hololens/}}, a holographic computer. 
The HoloLens is  built onto a headset for an augmented reality experience. 
It is a versatile and powerful edge-device that can work offline and also connect to the cloud.  
Microsoft aims to design standard computing, data analysis, medical imaging, and gaming-at-the-edge
tools using the HoloLens.

\item The Linux Foundation recently launched the LF Edge\footnote{\url{https://www.lfedge.org/}}
project to facilitate applications at the edge and establish a common open source framework that is independent of the operating systems and hardware.
EdgeX Foundry\footnote{\url{https://www.edgexfoundry.org/}} is another Linux Foundation project
that involves developing a framework for industrial IoT edge applications \cite{edgexfoundry}.

\item GE\footnote{\url{https://www.ge.com/digital/iiot-platform/predix-edge}},
IBM \cite{IBMEdge},
Cisco \cite{CiscoEdge},
VMWare,
and Dell\footnote{\url{https://www.dellemc.com/en-us/service-providers/edge-computing.htm}}
have all committed to investing in edge computing; 
 VMware is developing a framework for boosting enterprise IoT efforts at the edge \cite{VMwareEdge}. 
\end{itemize}

In the last few years, edge computing is being increasingly used for the {\it deployment of ML
  based intelligent systems in resource-constrained environments} \cite{vvedge19, DBLP:journals/corr/abs-1809-00343, voghoei2018deep, 8763885, 8736011},
which is the motivation for this survey.
A note on terminology: fog computing is a related term that describes an architecture where the `cloud is extended'
to be closer to the IoT end-devices, thereby improving latency and security by performing computations near the network edge \cite{Iorga2018FogCC}.
The main difference between fog and edge computing pertains to where the data is processed:
in edge computing, data is processed either directly on the devices to which the sensors are attached or on gateway devices physically very close to the sensors;
in the fog model, data is processed further away from the edge, on devices connected using a local area network (LAN) \cite{FogvsEdge}.

The use of ML models in edge environments  creates a distributed intelligence
architecture. In this respect, this survey describes how the
following problems are being addressed by the research community:

\begin{itemize}[topsep=0pt,itemsep=0ex,partopsep=1ex,parsep=1ex]
    \small
\item What are the practical applications for edge intelligence?
\item How have existing ML algorithms been adapted for deployment on the edge?
\item What is the state of the art in distributed training (and inference)?
\item What are the common hardware devices used to enable edge intelligence?
\item What is the nature of the emerging software ecosystem that supports this new end-edge-cloud
  architecture for real-time intelligent systems?
\end{itemize}  


\begin{table}[bt]
  \centering
  \scriptsize
  \caption{A summary of related surveys.}
  \label{table:existingSurvey}
  \renewcommand{\arraystretch}{2}
  \centering
  \begin{tabular}{|m{2.2cm}|m{5cm}|m{1cm}|m{1cm}|m{1cm}|m{1cm}|m{1cm}|} 
    \hline
    		{} & {} & \multicolumn{5}{c|}{Scope}\\\cline{3-7}
        {Paper} & \makecell{Summary} & {Tradit-ional ML} & {DL} & {Applic-ations} & {Soft-ware} & {Hard-ware}\\
        \hline
        {Park et al. \cite{park19Wireless}} & {Summarised the techniques that help enable ML at network edge} & \cmark & \cmark & \cmark & \xmark & \xmark 
        \\ 
        \hline
        {Zhou et al. \cite{8736011}} & {Reviewed the techniques/frameworks for training and inference of DL on edge devices} & \xmark & \cmark & \cmark & \cmark & \xmark \\
        \hline
        {Chen and Ran \cite{8763885}} & {Conducted a survey on techniques that help to speed-up DL training and inference on edge devices} & \xmark & \cmark & \cmark & \xmark & \xmark\\
        \hline
        {Lim et al. \cite{Lim2019FederatedLI}} & {Provided a comprehensive analysis of federated learning and how this technique helps deploy DL on edge settings} & \xmark & \cmark & \cmark & \cmark & \xmark\\ 
        \hline
        {Wang et al. \cite{wang2020convergence}} & {Reviewed the development of DL through edge computing architecture from a network \& communication perspective} & \xmark & \cmark & \cmark & \cmark & \cmark\\
        \hline
        {This survey} & {Analyzes ML techniques (both deep learning \& other classical methods) for resource-constrained edge settings.} & \cmark & \cmark & \cmark & \cmark & \cmark \\
        \hline
  \end{tabular}
\end{table}

As shown in \autoref{table:existingSurvey}, unlike ours, prior surveys about ML using edge computing 
(except \cite{park19Wireless}) do not cover classical ML techniques like random forests, Support-vector machine (SVM), etc. that have been used very frequently in edge computing based AI applications.
Instead, existing surveys have focused on deep learning (DL) and its applications, as shown in the table.
Note that Park et al. \cite{park19Wireless} focused on wireless communication and also
did not discuss hardware/software platforms for edge ML.
To summarize, the contributions of this paper include:

\begin{itemize}[topsep=0pt,itemsep=0ex,partopsep=1ex,parsep=1ex]
    \small
\item a survey of both Classical machine learning and deep learning from a resource-constrained edge computing perspective,
\item common software/hardware platforms used in DL and edge computing architectures, and
\item a detailed discussion of a wide range of applications pertaining to ML and edge computing
  from an operational perspective.
\end{itemize}
    
\subsection{Survey Structure and Methodology}
This survey focuses on machine learning systems deployed on edge-devices, and also covers efforts made to train ML models on edge-devices.
We begin by summarizing the ML algorithms that have been used in edge computing (\autoref{sec:MLAlgoForEdge}),
discuss ML systems that exploit edge-devices (\autoref{sec:EdgeMLApplications}),
and then briefly describe the frameworks, software and hardware
used to build and deploy intelligent systems at the edge (\autoref{sec:framewswhw}).  
The challenges hindering more widespread adoption of mobile edge computing
and possible directions of future research (\autoref{sec:ChallengesAndFD}) are also discussed.

\begin{figure*}[t]
\center
  \fbox{\includegraphics[width=0.8\textwidth]{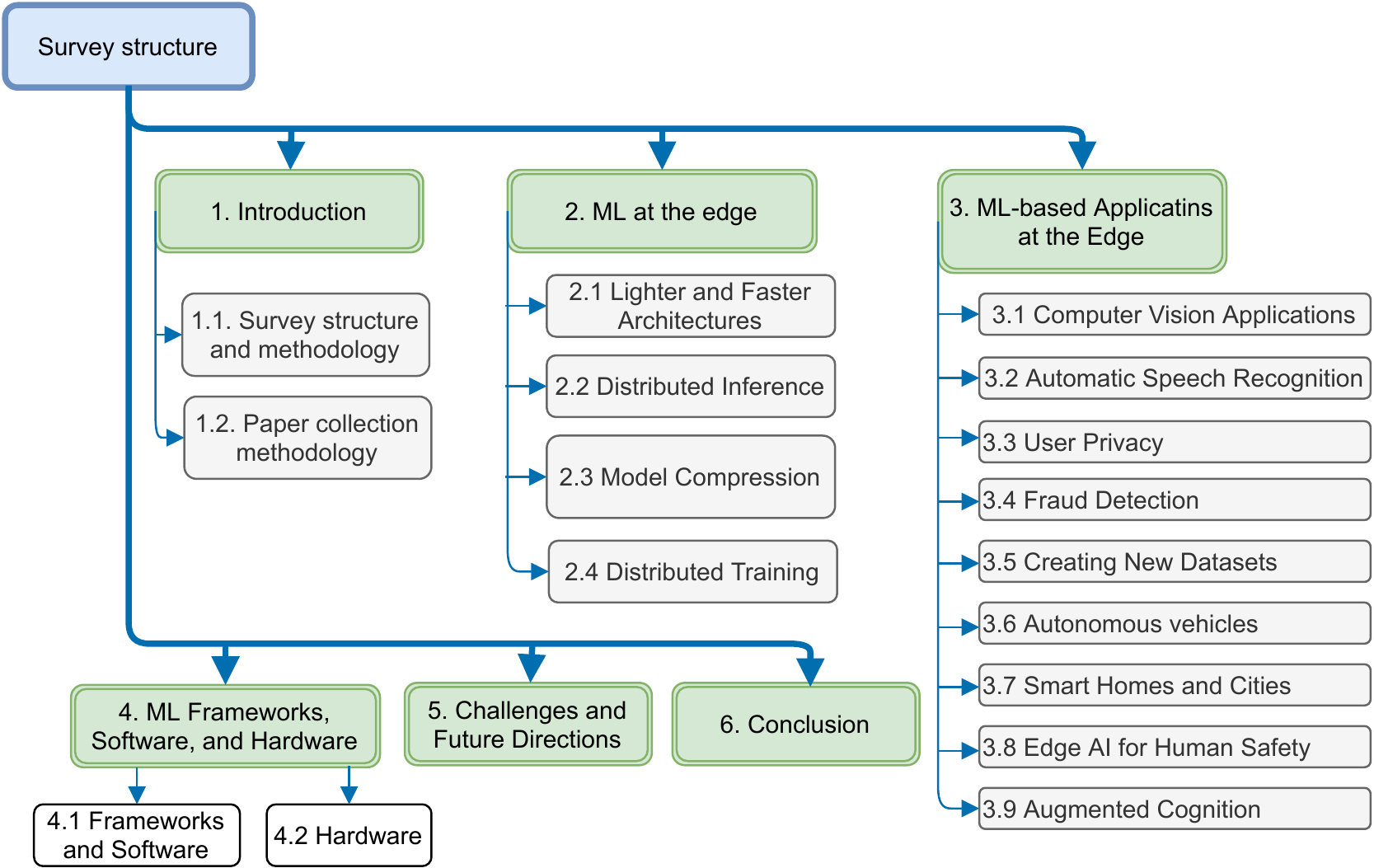}}
  \caption{An overview of the structure of the survey.}
  \label{fig:structure}
\end{figure*}

\subsection{Paper Collection Methodology}
We conducted exact keywords searches on Google Scholar, Microsoft Academic, DBLP, IEEE, ACM digital library and
arXiv to collect papers related to ML and edge computing, resulting in 124 papers. 
The following search terms were used:

\begin{itemize}[topsep=0pt,itemsep=-1ex,partopsep=1ex,parsep=1ex]
    \small
\item (machine | deep) learning + edge computing
\item (machine | deep) learning + (resource-constrained devices | IoT)
\item modified learning algorithms + edge computing
\item (SVM | k-means | decision trees | convolutional neural networks | recurrent neural networks) + edge computing
\item (compressed | sparse | acceleration | deep learning | machine learning) + edge computing 
\item resource-efficient deep neural networks + edge computing
\end{itemize}

\noindent The following questions were used to determine whether to include a paper in the survey:

\begin{itemize}[topsep=0pt,itemsep=-1ex,partopsep=1ex,parsep=1ex]
    \small
\item Was edge computing used to improve ML efficiency in a networked ML system?
\item Were ML algorithms specialized for resource-constrained devices designed or applied in a new setting?
\item Were new results describing the deployment of ML systems on the edge reported?
\end{itemize}	

\noindent If any of the questions above could be answered in the affirmative, the paper was included.
After careful analysis regarding the three questions mentioned above,
we considered  88 out of the 124 papers collected for this survey.
Papers published before May 2020 were considered for this survey.

\section{MACHINE LEARNING AT THE EDGE: TRAINING, COMPRESSION, AND DEPLOYMENT} \label{sec:MLAlgoForEdge}
Modern real-world applications,  such as autonomous cars and mobile robots,
increasingly require fast and accurate automated decision making capabilities.
Large computing devices can provide the needed real-time computational power, but in many
situations, it is nearly impossible to deploy them without significantly affecting performance, e.g. on a robot.
Instead, ML models must be adapted in a way that makes them suitable for deployment on (small) edge-devices
that have limited computational power and storage capacity.
This section describes Classical ML and deep learning algorithms that have been used in resource-constrained settings.

           

With the increasing amount of information being generated by various kinds of devices placed at the network edge,
the demand for ML models that can be deployed on such resource-constrained devices has also increased.
Training machine learning models, especially deep learning model, typically requires a lot of computational power and a large number of training examples.
Low-power IoT devices, such as on-board cameras, are continuous sources of data but their limited storage and compute capabilities
make them unsuitable for training and inference of DL models.
In systems with a  end-edge-cloud hierarchy, centralized training is a problem due to limited bandwidth available. To address all these issues, edge-servers are increasingly being placed near these IoT end-devices and used for deploying DL models that operate on IoT-generated data. Efforts have been made for distributed training by using the edge-devices only (\autoref{sec:distTraining}).

Due to the limited memory and compute power of edge-devices, it is important to tailor the model so that it can fit
and execute efficiently on those devices. Therefore, techniques have also been developed for compression ML models to make the models lighter and faster and to aid their deployment on the edge (\autoref{subsec:storagelim}). \autoref{table:majorDLcontribution} presents a summary of the major contribution in ML algorithms at the network edge.


\begin{table}[t]
  \centering
  \scriptsize
  \caption{Major contributions in machine learning at the edge. }
  \label{table:majorDLcontribution}
  \renewcommand{\arraystretch}{1.4}
  \begin{tabular}{|p{3cm}|c|p{7cm}|} 
    \hline
        {Challenges in Edge DL} & {Paper} & {Key contribution}\\
        \hline
            {~} &  Zhang et al.~\cite{Zhang18ShuffleNet} &  ShuffleNet: efficient CNN models for mobile and embedded devices\\\cline{2-3}
           {~} & Howard et al.~\cite{DBLP:journals/corr/HowardZCKWWAA17} &  MobileNets:      Small, efficient, low-power, low-computation and low-latency CNN models\\\cline{2-3}
           {Lighter and Faster Architectures (\autoref{sec:lightFasterArchs})} & Tan and Le~\cite{tan2019efficientnet} & Efficient scaling of CNN architecture while retaining accuracy\\\cline{2-3}
           {~} & Tang et al.~\cite{tang2020understanding} & Address overparameterization during training to improve model quality\\\cline{2-3}
           {~} & Kusupati et al.~\cite{Kusupati18FastRNN} & Addressed unstable training \& inefficient prediction,
           while keeping maintaining accuracies and small model size\\\hline
           {~} & Wang et. al~\cite{Wang2018WhenEM} & Distributed training for gradient-descent based algorithms.\\\cline{2-3}
           {~} & Nishio and Yonetani~\cite{Nishio2018ClientSF} &  FedCS: faster learning federated learning for model training\\\cline{2-3}
            {~} & Lui et al.~\cite{Liu2019EdgeAssistedHF} & Intermediate edge aggregation to reduce communication cost during federated learning\\\cline{2-3}
            {Distributed Training and high network communication cost (\autoref{sec:distTraining})} & Wang et al.~\cite{wang19CMFL} & Local \& global update comparison technique in federate learning  to reduce communication cost\\\cline{2-3}                     
            {~} & Tao and Li~\cite{Tao18eSGD} & Developed eSGD that is capable of reducing the gradient size of a CNN model by up to 90\%\\\cline{2-3}
            {~} & Lin et al.~\cite{Lin17DSGD} & Deep gradient compression to reduce gradient up to 600 times
            maintaining accuracy\\
\hline
{~} & Gupta et. al~\cite{pmlr-v70-gupta17a} & ProtoNN, an algorithmic technique for classification on resource-constrained devices.\\\cline{2-3}
            {~} & Kumar et al.~\cite{bonsai2kedge} & Bonsai, a tree-based algorithm for ML prediction on resource-constrained IoT devices.\\\cline{2-3}
{~} & Pradeep et al.~\cite{Pradeep18EdgeNet} & Quantization for CNN  deployment on embedded FPGAs\\\cline{2-3}
           {Weight Quantization and Model Compression (\autoref{sec:compDL})} & Gupta et al.~\cite{Gupta:2015:DLL} & Reduced memory footprint using 16-bit fixed-point for weights\\\cline{2-3}
           {~} & Ogden and Guo~\cite{ogden18MODI} & Used 8-bit integers to represent CNN weights and reduced model size by 75\%\\\cline{2-3}
           {~} & Iandola et al.~\cite{Iandola16SqueezeNet} & SqueezeNet: an architecture
           with 50 times fewer parameters than AlexNet\\
\hline
            {~} & Ogden and Guo~\cite{ogden18MODI} & Developed a distributed platform for mobile deep inference\\\cline{2-3}
            {~} & Teerapittayanon et al.~\cite{7979979} & Mapped DNNs across distributed computing hierarchies for faster inference\\\cline{2-3}
            {~} & Mao et al.~\cite{Mao7927211} & MoDNN: a distributed mobile computing system for deploying DNNs\\ \cline{2-3} 
           {~} & Hadidi et al.~\cite{10.1145/3316781.3322474} & Partitioning of DNN computations across devices and for reduced latency\\\cline{2-3} 
           {Distributed deep learnings inference (\autoref{sec:distrInference})} & Hadidi et al.~\cite{8411096} & Dynamic distribution of DNN computations across devices\\\cline{2-3}
           {~} & Huai et al.~\cite{9060297} & Faster DNN inference by partitioning the model across multiple devices\\
\cline{2-3}
		{~} & Stahl et al.~\cite{Stahl19DDNN} & Layer fusion scheme to reduce communication time during DNN inference \\
		\hline
           
  \end{tabular}
\end{table}



\subsection{Lighter and Faster Architectures} \label{sec:lightFasterArchs} 
In order to enable the deployment of deep learning models on resource-constrained edge-devices,
the following goals need to be met while ensuring minimal loss in accuracy:
\begin{itemize}
    \small
\item the model size needs to be kept small by using as few trainable parameters as possible and
\item the inference time and energy needs to be kept low by minimizing the number of computations.
\end{itemize}

\subsubsection{\textbf{Convolutional neural networks}} \label{sec:DSC}
It is challenging to deploy CNNs on edge-devices, owing to their limited memory and compute power.
\emph{Depthwise separable convolutions}  provide a lightweight CNN architecture (see \cite[\S 3.1]{DBLP:journals/corr/HowardZCKWWAA17}) 
by replacing the sum of multiple convolutional operations
by a weighted sum of results obtained from only one convolutional operation. 
A standard CNN model uses each convolutional layer to generate a new set of outputs by filtering and summing the input channels.
In contrast, depthwise separable convolutions divide each convolutional layer into two separate layers which serve the same purpose as a single convolutional layer while also greatly reducing the model size and computational cost during inference.
For a convolutional filter of size $K \times K$ operating on an input volume of $M$ channels and
producing an output of $N$ channels,
a depthwise separable convolution results in the computational cost being reduced
by a factor of $1/N + 1/K^2$
and the number of parameters being reduced by a factor of $1/N + (M/(K^2*M+1))$, 
compared to the standard convolution.
For example, if  $K = 5$, $M = 64$, and $N = 128$, depthwise separable convolutions
will result in computations as well as the number of parameters being reduced by approximately 21 times.

Depthwise separable convolutions were exploited in MobileNets \cite{DBLP:journals/corr/HowardZCKWWAA17} 
to reduce the number of parameters from 29.3 million to 4.2 million and the number of computations
by a factor of 8 while containing the accuracy loss to about only $1\%$.
ShuffleNet  \cite{Zhang18ShuffleNet} is another architecture that uses two other types of operations
(viz. group convolutions and channel shuffling)
along with depthwise separable convolutions
to produce models that are lighter and faster than MobileNet while also being more accurate.

EfficientNet provides a  method for systematically scaling up CNN models in terms of network depth, width,
and input image resolution \cite{tan2019efficientnet}.
On ImageNet, it produced models yielding state-of-the-art accuracy using 8 times lesser memory and having 6 times
faster inference.
It also achieved  state-of-the-art transfer learning performance using an order-of-magnitude fewer parameters.
However, the inference speed of EfficientNet on resource-constrained edge-devices does not scale well.
This is addressed by EfficientNetLite\footnote{\url{https://github.com/tensorflow/tpu/tree/master/models/official/efficientnet/lite}}, a suite of EfficientNet models tailored for edge-devices.
Users can choose from a range of small, low latency models to relatively larger, high accuracy models.
For ImageNet, even their largest model performed  real-time image classification with state-of-the-art
 on a smart-phone CPU.

Other techniques for model size reduction include knowledge distillation
which refers to training a small model to mimic the outputs of a larger,
more complex trained model \cite{tang2020understanding, Sharma2018EKT}
and low-rank factorization which allows the decomposition of convolutional operations in order to reduce parameters
and speed-up computations during CNN operations \cite{cheng2017survey}.

\subsubsection{\textbf{Recurrent neural networks}}
Recurrent Neural Networks (RNNs) are powerful neural networks used for processing sequential data, but suffer from
unstable training and inefficient prediction.
Techniques to address these issues, such as unitary RNNs and gated RNNs,
increase the model size as they add extra parameters. 
FastRNN and FastGRNN \cite{Kusupati18FastRNN} are new architectures that provide stable training and good accuracies
while keeping model size small.
In FastRNN, an additional residual connection is used that has only two additional parameters to stabilize the training by generating well-conditioned gradients.

Gated RNNs are typically more expressive than ungated RNNs but require additional parameters.
FastGRNN is a gated version of FastRNN that does not increase the number of parameters.
Model size was reduced by quantizing weights to be integer valued only.
Prediction time was reduced by using piecewise-linear approximations to ensure that all computations use integer arithmetic only.

FastGRNN has 2-4 times fewer parameters than other leading gated RNN models such as LSTM, GRU, but provides the same or sometimes better accuracy than gated RNN models \cite{Kusupati18FastRNN}.
It is possible to fit FastGRNN in 1-6 kilobytes which makes this algorithm suitable for IoT devices, such as Arduino Uno.
Since competing RNN models cannot fit on an Arduino Uno, the authors  compared FastRNN and FastGRNN with other models
on an Arduino MKR1000 and reported 18--42 times faster prediction.
FastRNN and FastGRNN  are open-source \cite{MSEdgeML} and
have been benchmarked on a number of problems including utterance detection, human activity recognition, and language modeling.

\subsection{Distributed Training} \label{sec:distTraining}
Distributed training of DL models on multiple edge-devices and/or the cloud has two significant benefits:
\begin{itemize}
    \small
\item Data acquired from end-devices can be used for training/re-training at the edge-servers,
thereby reducing the communication overhead of transmitting to and training on a central cloud server.
\item It also leads to greater privacy since no central location has access to data produced by end-devices.
\end{itemize}
The next two sections discuss  collaborative and privacy-aware learning on the end-edge-cloud architecture.\\

\subsubsection{Distributed training for algorithms based on gradient descent}
Wang et al. introduced a technique to \emph{train ML models at the edge without the help of external computation systems}
such as cloud servers \cite{Wang2018WhenEM}.
They focused on algorithms that use gradient-based approaches for training 
(including SVMs, K-means, linear regression and convolution neural networks (CNNs) ).
Their technique minimizes the loss function of a learning model by using the computational power of edge-devices only. 

Local gradient descent is performed on multiple edge-devices based on data received at each device (i.e. on local datasets).
This produces locally updated copies of the ML model.
These copies are sent to another edge-device, known as the aggregator, that computes a weighted average of all the locally updated models.
This averaged model is sent back to all the edge-devices so that each device now carries a model averaged over training data from all edge-devices.
The next iteration performs local updates once again and the process repeats until resource
consumption reaches a prespecified budget.

To demonstrate the effectiveness of this technique, three Raspberry Pi devices  and
a laptop computer were used as edge-devices. They were connected via WiFi
to an additional laptop computer that was used as the aggregator. 
Three datasets (viz., MNIST, Facebook metrics, and User Knowledge Modeling) were used
to evaluate the ML models. 
The technique's performance was close to the optimum
for training different ML models on different datasets.

\subsubsection{Federated learning}
\begin{figure}[ht]
    \centering
   \fbox{\includegraphics[width=0.45\textwidth]{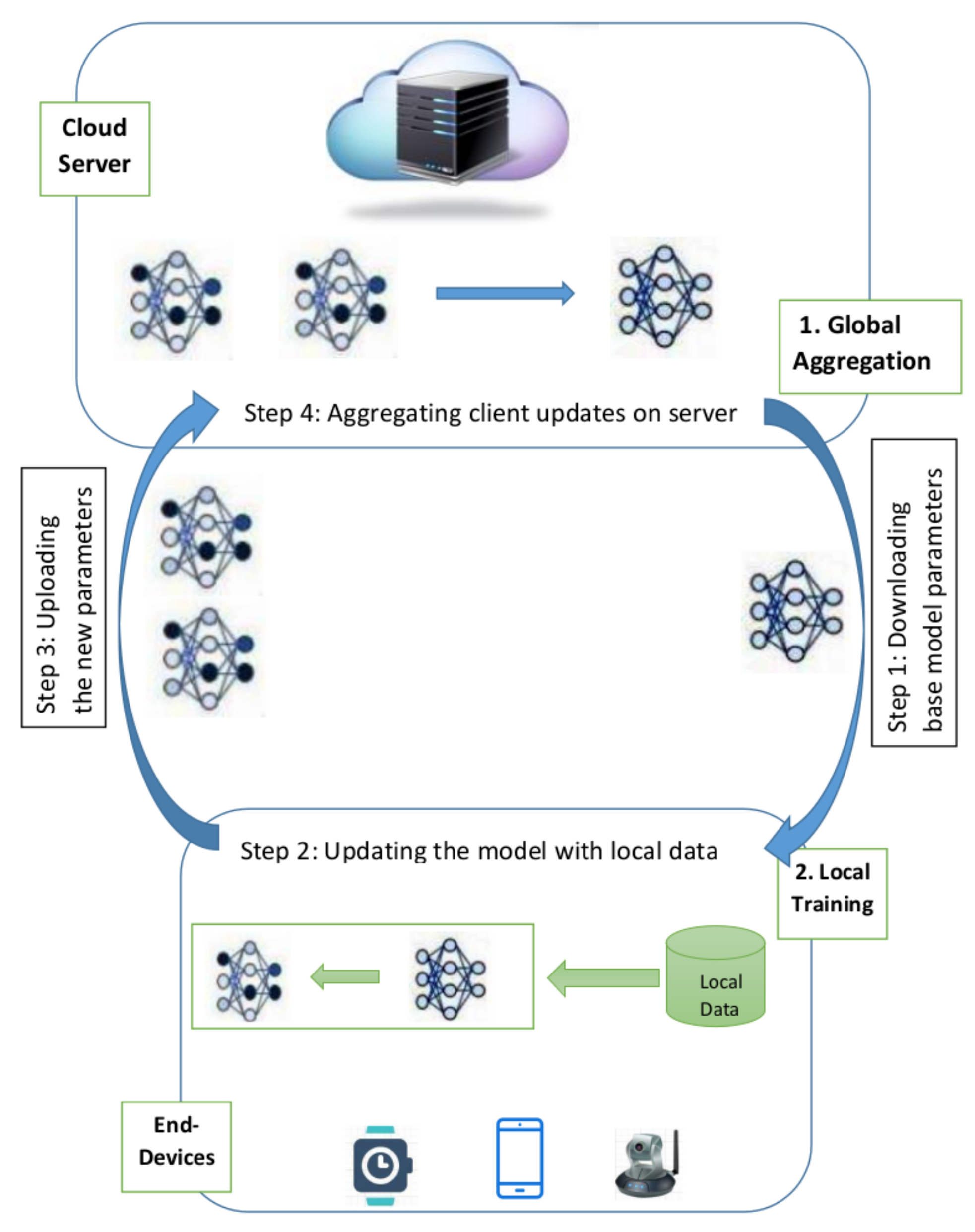}}
   \caption{Federated learning allows training on end-devices where the data is produced. First, end-devices download parameters of a trainable ML model from the cloud server. Then, those devices update the model locally with their own data. After that, all end-devices upload the updated model parameters. Finally, the cloud server aggregates multiple client updates to improve the model.}
         \label{fig:flearning}
  \end{figure}

Federated Learning (FL) is a family of distributed ML methods that involve
collaborative training of shared prediction DNN models on devices
such as mobile phones \cite{McMahan2016CommunicationEfficientLO}.
All training data is kept on the end-devices 
and model training occurs in powerful local or cloud computing infrastructure.
 In general, there are two steps in the FL training process, viz. 
\begin{enumerate*}[label=\itshape\roman*\upshape)]
\item local training and 
\item global aggregation.  
\end{enumerate*}
In local training, the end-device downloads the model from a central cloud server,  computes an updated model using that local data to improve model performance. 
After that, an encrypted communication service is used to send a summary of all updates made by the end-device to the server. 
The server aggregates these updated models (typically by averaging) to construct an improved global model,  as illustrated in \autoref{fig:flearning}.
This decentralized ML approach ensures the maximum use of available end-devices and does not share any data among end-devices, which helps to enhance the security and privacy of the local data. However, federated learning faces challenges that include communication overhead, interoperability of heterogeneous devices, and resource allocation \cite{Lim2019FederatedLI, li2019federated}.

Federated learning with heterogeneous edge clients can lead to slow learning when some edge clients have very low-resources.
 To address this problem, Nishio and Yonetani introduce FedCS which performs federated learning with an
intelligent client selection mechanism to ensure faster learning of high performing ML models \cite{Nishio2018ClientSF}.


Exchanging model parameters and other data between edge-devices and cloud servers is mandatory
for training an edge-cloud-based DL model.
However, as the size of the training model increases, more data needs to be exchanged between edge-devices and cloud servers.
The high network communication cost is a bottleneck training a model.
Various techniques have been  proposed to reduce communication costs during training:

\begin{itemize}
    \small
\item Lui et al. introduced intermediate edge aggregation before FL server aggregation \cite{Liu2019EdgeAssistedHF}.

\item  Wang et al. compare each client's local model update with the global update of the model at a central server. If the local update differs significantly from the global update, it is not communicated to the central server since it will become less relevant when updates are aggregated \cite{wang19CMFL}.  

\item Tao and Li introduced a new method called 
Edge Stochastic Gradient Descent (eSGD) \cite{Tao18eSGD} 
that is capable of reducing the gradient size of a CNN model by up to 90\%
by communicating only the most important gradients.
On MNIST, they obtained an accuracy of 91.22\% with a 50\% gradient drop.

\item Lin et al. studied the effect of gradient exchange in distributed stochastic gradient descent (DSGD)
and found 99.9\% of gradient exchanges to be redundant \cite{Lin17DSGD}.
This observation inspired them to propose \emph{deep gradient compression},
which compress the gradient from 270 to 600 times without losing much accuracy.
They applied momentum correction, local gradient clipping, momentum factor masking,
and warm-up training methods   to reduce the size of gradient updates for training ResNet-50 from 97 MB to 0.35 MB,
  and for training DeepSpeech  \cite{hannun2014deep} from 488 MB to 0.74 MB. 
\end{itemize}


\subsubsection{Privacy in learning}
Privacy of sensitive training data becomes an issue when training is performed in a distributed fashion.
Techniques derived from cryptography can be applied for hiding updates of local devices
from a central aggregation server \cite{bonawitz2017practical} and also for hiding the aggregated update
from the local devices \cite{DBLP:journals/corr/abs-1712-07557}.
While ensuring such local and global privacy mechanism, it is important to
\begin{enumerate*}[label=\roman*)]
\item contain any drop in accuracy
\item ensure low overheads of computation and communication and
\item introduce robustness to communication errors and delays.
\end{enumerate*}

Mao et al. presented a privacy-aware DNN training architecture that uses differential privacy  to protect user data during the training  \cite{Mao18DLPrivacy}.
This deep learning scheme is capable of training a model using multiple mobile devices and the cloud server collaboratively, with minimal additional cost. First, this algorithm selects one convolutional layer to partition the neural network into two parts. One part runs on edge-devices, while another part runs on the server (AWS cloud server). The first part takes raw data with sensitive user information as input and uses a differentially private activation algorithm to generate the volume of activations as output. These output activations contain Gaussian noise, which prevents external cloud servers from reproducing original input information using reversing activations. These noisy output activations are then transmitted to cloud servers for further processes. The servers take output activations as input and run the subsequent training process. This model was evaluated using a Nexus 6P phone and AWS based servers. The authors report that they have achieved good accuracy by deploying this model on Labeled Faces in the Wild dataset (LFW). 

In order to avoid centralized aggregation of potentially sensitive information,
blockchains have been explored as an alternative \cite{rahman2018blockchain, 10.1145/3285017.3285027}.
However, the compute power and energy requirements of mining on edge-devices remains an important
challenge \cite{wang2020convergence}.

\subsection{Model Compression} \label{sec:compDL}

In this section, we discuss compressed ML techniques designed for memory-scarce IoT devices.
  In the literature, we found some compression techniques for classical ML used in resource-constrained settings,
  which may also open up future research in model compression area.
  A popular technique for classical ML compression is to project data to a lower dimensional space and learn in the projected space. However, learning in a low-dimensional data space leads to poor accuracy.
  Techniques for regaining accuracy are covered during the discussion on model compression ({\autoref{subsec:storagelim}}).

In order to run on low-memory devices, deep learning models need to be reduced in size
while containing the loss in accuracy.
There are two broad approaches for doing this
\begin{enumerate*}[label=\itshape\alph*\upshape)]
\item  \emph{quantization}, which reduces the precision with which parameter values are stored
which in turn lowers the memory footprint and therefore potentially makes computations faster, and,
\item \emph{model pruning}, which reduces the number of model parameters and therefore improves storage and compute time.
\end{enumerate*}
Note that both methods can be applied on a given model one after the other.
Commonly used DL model quantization and pruning techniques are discussed in {\autoref{subsub:modelquantizedandpruned}}.
A detailed description of model compression techniques can also be found in the surveys by Cheng et al. \cite{cheng2017survey}
and Choudhary et al. \cite{choudhary2020comprehensive}.

\subsubsection{Overcoming storage limitations with compression}\label{subsec:storagelim}
{\em ProtoNN} is a new technique, designed by Gupta et al., for \emph{training an ML model on an edge-device}
that can perform real-time prediction tasks accurately \cite{pmlr-v70-gupta17a}.
It is a k-nearest neighbor (k-NN) based algorithm with low storage requirements. 
The following issues arise when trying to implement a standard k-NN on a
resource-constrained edge-device: 

\begin{itemize}[topsep=0pt,itemsep=-1ex,partopsep=1ex,parsep=1ex]
    \small
\item it requires the entire training dataset  to be stored, 
  which is difficult on resource-constrained devices

\item the distance of a given  sample from each training example needs to be computed,
  which inhibits real-time
  prediction on resource-constrained devices due to their limited computational power

\item the results are sensitive to the choice of the distance metric.
\end{itemize}

To address these issues, ProtoNN projects data to a lower dimensional space
using a sparse-projection matrix and
selects only a few prototype vectors/samples from this low-dimensional data.
The algorithm learns the projection matrix and the prototypes and their  labels 
in a joint optimization step.
The joint learning process helps to compensate for the reduction in accuracy
caused due to the projection into lower dimensional space and selecting only
a few representative prototypes.

For settings needing  extremely small  models ($<2$ kB),
ProtoNN was shown to outperform the baseline compressed models like GBDT, RBF-SVM, 1-hidden layer NN, etc in character recognition dataset, 
while in settings allowing 16-32 kB memory, it matched the performance of the baseline compressed models.
This was true for binary, multiclass as well as multilabel datasets.
Compared to the best uncompressed models, ProtoNN was only 1-2\% less accurate
while consuming 1-2 orders of magnitude less memory (in multiclass and multilabel settings).
On severely resource-constrained IoT devices, energy usage
and time to make predictions are important considerations \cite{yazici2018edge}.
On an Arduino Uno (2 kB RAM),
ProtoNN was shown to be more efficient in both aspects compared with existing methods.
ProtoNN is available as part of Microsoft's EdgeML library  \cite{MSEdgeML}
and Embedded Learning Library \cite{msell}.

Tree-based ML algorithms are commonly used for classification, regression and ranking problems.
Even though the time-complexity of tree-based algorithms is logarithmic with respect to the size of the training data, their space complexity is linear,
so they are not easily amenable to deployment on resource-constrained devices.
Aggressively pruning or learning shallow trees are ways to shrink the model but lead to poor prediction results.
{\em Bonsai} is a novel tree-based algorithm developed by Kumar et al. that  significantly outperforms state-of-the-art techniques
in terms of model size, accuracy, speed, and energy consumption \cite{bonsai2kedge}.

To reduce the model size, Bonsai learns a single decision tree that is shallow as well as sparse.
These three decisions should naturally decrease overall accuracy. Bonsai manages to retain high accuracy by:

\begin{itemize}[topsep=0pt,partopsep=1ex,parsep=1ex]
    \small
\item
 performing a low-dimensional linear projection of the input.
  The linear projection can be performed in a streaming fashion, i.e. without storing the whole input in RAM. This is important for solving large input size problems in low-memory environments.
This low-dimensional projection is learned jointly with all the other parameters of the model. Such joint learning leads to higher accuracy.  

\item making a shallow decision tree more powerful (non-linear) by:
\begin{enumerate*}[label=\itshape\roman*\upshape)]
\item using non-axis-aligned branching,
\item enabling all internal nodes to become classifiers as well, and
\item making the final classification result a sum of all results along the path traversed by the input vector.
\end{enumerate*}
\end{itemize}

Bonsai has been tested with a number of binary and multi-class datasets.
When deployed on an Arduino Uno, Bonsai required only 70 bytes for a binary classification model and 500 bytes a 62-class classification model. Its predication accuracy was up to $30\%$ higher than other resource-constrained models and even comparable with unconstrained models. Prediction times and energy usage were also measured and shown to be better for Bonsai models.
Bonsai is available as part of the EdgeML library  \cite{MSEdgeML}. 

\subsubsection{Model quantization, and pruning} \label{subsub:modelquantizedandpruned}
SqueezeNet \cite{Iandola16SqueezeNet} is a parameter-efficient neural network used in resource-constrained settings.
This small CNN-like architecture has 50 times fewer parameters than AlexNet while preserving
AlexNet-level accuracy on the ImageNet dataset.
It reduced model size by 
\begin{enumerate*}[label=\roman*)]
\item replacing $3 \times 3$ convolutions with $1 \times 1$ convolutions,
\item  using weighted sums of input channels via $1\times1$ convolutions to reduce the number of input channels,
\item  converting input channels to $3\times 3$ filters only, and
\item  using delayed downsampling to achieve higher classification accuracy.
\end{enumerate*}
By using Deep Compression (weight pruning + quantization + Huffman encoding) \cite{han2015deep},
SqueezeNet was further compressed to 0.47 MB, which is 510 times smaller than 32-bit AlexNet.
These techniques decrease the number of parameters at the cost of accuracy.
To compensate, the authors downsample later in the network to have larger activation maps which lead to higher accuracy.
The authors report that SqueezeNet matches or exceeds the top-1 and top-5 accuracy of AlexNet while using a 50 times smaller model.

Pradeep et al. deployed a CNN model on an embedded FPGA platform \cite{Pradeep18EdgeNet}.
They use low bit floating-point representation\footnote{8-bit for storage and 12-bit during computations.}
to reduce the computational resources required for running a CNN model on FPGA.
Their architecture was tested with SqueezeNet on the ImageNet dataset.
The authors reported having achieved $51\%$ top-1 accuracy
on a DE-10 board at 100 MHz that only consumed 2W power.
Gupta et al. used 16-bit fixed-point representation in stochastic rounding based CNN training \cite{Gupta:2015:DLL}.
This lower bit representation significantly reduced memory usage with little loss in classification accuracy.
Ogden and Guo use grouped weight rounding and 8-bit quantization to reduce model size by $75\%$
and increase inference speed while containing the accuracy drop to approximately $6\%$ \cite{ogden18MODI}.

A few sensitive regions of a feature map have greater impact on deep learning inference compared to other
  regions \mbox{\cite{9138970}}.
  Conventional quantization techniques are applied to entire network layers or on kernel weights without considering the dynamic properties of the feature map. Such quantization sometimes hurt the overall performance of a deep learning model despite large reductions of energy usage. Song et al. developed \mbox{\em{DRQ}}, a dynamic region-based quantization that can detect the sensitive regions in the feature map dynamically \mbox{\cite{9138970}}. To preserve the accuracy, DRQ implies high-fidelity quantization on the sensitive regions and low-fidelity quantization on the insensitive regions.
  This system achieved an improvement of  $3\%$ in the prediction
accuracy improvement compared to the state-of-the-art mixed-precision
quantization accelerator \mbox{\em{OLAccel}} \mbox{\cite{10.1109/ISCA.2018.00063}}.

The Sparse CNN (SCNN) accelerator compresses convolutional neural networks by detecting and pruning the zero-valued activations from the ReLU operator and the zero-valued weights of the network during training \mbox{\cite{10.1145/3140659.3080254}}.
  However, traditional weight pruning techniques sometimes incur additional storage overhead and reduce performance accuracy \mbox{\cite{10.1145/3079856.3080215}}.
  \mbox{\em{Scalpel}}, a customizing DNN pruning technique, overcomes these weaknesses by introducing two new methods:  SIMD-aware weight pruning and node pruning.
  These pruning techniques were tested on microcontrollers and CPUs and achieved
  a reduction in model size of  $88\%$ and $82\%$ (resp.) and a  geometric mean performance speed-up of $3.54$ times and $2.61$ times (resp.) \mbox{\cite{10.1145/3079856.3080215}}.

\mbox{\em{Eager Pruning}} is a model pruning technique that observes the rank of the significant weights of the DNN
  and prunes insignificant weights during training \mbox{\cite{10.1145/3307650.3322263}}.
  This method was shown to reduce training computation of a network by up to $40\%$ while still maintaining the original accuracy.
  Eager pruning applied on eight different deep learning architectures
  achieved an average speedup of $1.91$ times over a state-of-the-art hardware accelerator.

Yang et al. developed a network compression method for the deployment of DNNs on memristor devices such as resistive random access memory (ReRAM) \mbox{\cite{10.1145/3307650.3322271}}.
  They built a \mbox{\em{Sparse ReRAM Engine}} which exploits activation and weight sparsity and prunes zero weights and zero activations.  When classifying objects on the ImageNet dataset using the VGG-16 CNN model, this pruning helps to compress the model and deploy it on ReRAMs achieving a speedup up to $42.3$ times and energy savings of up to $95.4\%$ over the state-of-the-art that does not exploit sparsity.

\subsection{Distributed deep learnings inference} \label{sec:distrInference}
Distributed deep neural network architectures allow the distribution of deep neural networks (DNNs) on the edge-cloud infrastructure in order to 
 facilitate local and fast inference on edge-devices wherever possible.
The success of a distributed neural network model depends on keeping inter-device communication costs as low as possible.
Several attempts have been made to split and distribute a model between edge-devices,
resulting in faster inference. 
DNN inference can be distributed in two ways, 
\begin{enumerate*}[label=\itshape\alph*\upshape)]
 \item vertically, along the end-edge-cloud architecture, and
 \item  horizontally, along multiple devices at the same level in the architecture.
\end{enumerate*}

\begin{figure}[h]
  \centering
  \fbox{\includegraphics[width=0.4\linewidth]{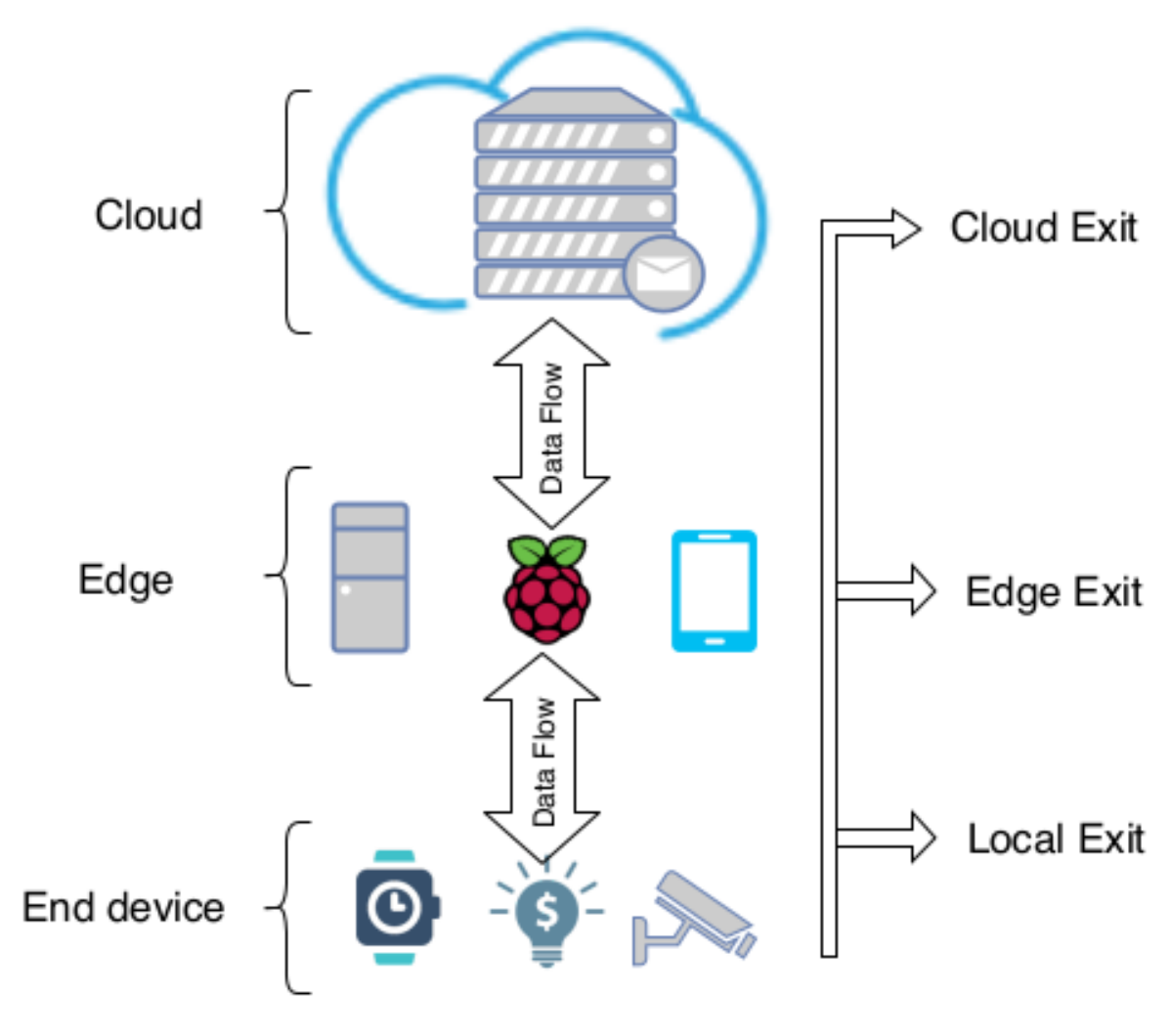}} 
  \caption{\small A high-level view of the distributed deep neural network (DDNN) approach
    designed by Teerapittayanon et al. \cite{7979979}.
    End-devices send summary information to a local aggregator which serves as a DNN layer. 
The  DDNN is jointly trained with all resource-constrained end-devices and exit points, so the network is capable of automatically collecting and combining input data from different end-devices. 
If the information collected from a particular end-device is sufficient to classify a sample then classification is done locally (i.e. on the end-device itself).
Otherwise, the information is sent to the edge-servers for further processing. 
If edge-servers can complete the classification, they send the result back to the end-devices. 
Otherwise, edge-servers send the information to the cloud, where the classification
process is completed, and the results returned to the end-devices.}
  \label{fig:DDNN}
 \end{figure}

\subsubsection{{\bf Vertically-distributed inference}}
We now describe two orthogonal approaches for exploiting the vertical direction of the end-edge-cloud architecture.
In the first approach mobile deep inference (MODI), the dynamic status of the architecture is considered to make optimal decisions about which model to run and where to run it.
In the second approach Early exit of inference (EEoI), resources along the vertical direction are exploited to exit
as early as possible in order to make optimal inferences while balancing the trade-off
between accuracy on the one hand and computational effort and latency on the other.\\

\noindent {\textbf{Mobile Deep Inference (MODI) Platform}}\\
Due to variable network conditions, inference on the end-edge-cloud architecture is a dynamic problem.
Static approaches such as on-device inference only or remote inference only are both sub-optimal.
Ogden and Guo \cite{ogden18MODI} propose a dynamic solution to this problem by designing a distributed platform for
mobile deep inference.
They propose to store multiple DL models (compressed and uncompressed)
in a centralized model manager and dynamically deciding
which model to run on which device based on the inference environment (memory, bandwidth, and power).
If the inference environment is resource-constrained, one of the compressed models is used, otherwise an
uncompressed model gets deployed, providing higher accuracy.
They show that on-device inference can provide up to 2.4 times speedup.
They also propose offloading remote inference to edge-servers when networks are slow
MODI runs jointly on end-devices and servers to provide the most suitable model for mobile devices based on device resources and installed applications.\\

\noindent {\textbf{Early exit of inference (EEoI)}}\\
Teerapittayanon et al. introduced distributed deep neural networks (DDNN\footnote{DDNN is open-source: \url{https://github.com/kunglab/ddnn}}),
where sections of a deep neural network are mapped across distributed computing hierarchies \cite{7979979}.
\autoref{fig:DDNN} shows the general structure of a distributed deep neural network.
Their method leverages cloud servers, resource-constrained edge-server, and end-devices
such as surveillance cameras for inference.
In such as model, a big portion of the raw data generated by sensors is processed on edge-devices
and then sent to the cloud for further processing.
This reduces their data communication cost.
The DNN receives input from end-devices and
produces final output in the cloud.
That is, the DNN is distributed over the entire end-edge-cloud architecture.
Additional classifiers are added to this network at the levels of end-devices and edge-servers
which allows a strategy for early exit at these classifiers to be used.
This allows for fast, early inference if a sample can be classified accurately by the lower classifiers alone.
Otherwise, the features extracted so far are propagated further up the network.
The whole DNN is trained jointly in an end-to-end fashion so that lower-level classifiers learn
to produce features that are also useful for subsequent classifiers.
The joint training takes place on a single, powerful server or on the cloud.
They also use binary neural networks \cite{hubara2016binarized, McDanel17BinarizedNN}
to reduce model size and increase inference speed.
Aggregation methods are used to fuse information from different end-devices.
This method of aggregation is extended to the edge-servers as well.
Using this technique reduced their data communication cost by a factor of more than 20 for a classification task
on a  multi-view multi-camera dataset.

\subsubsection{{\bf Horizontally-distributed inference}}
Mao et al. proposed a local distributed mobile computing system (MoDNN)
to deploy DNNs on resource-constrained devices \cite{Mao7927211}.
MoDNN uses a pre-trained DNN model and scans each layer of a DNN model to identify layer types.
If a convolutional layer is detected, the layer input is partitioned by a method called Biased One-Dimensional Partition (BODP).
BODP helps reduce computing cost by reducing the input size of convolutional layers.
If a fully connected layer is detected, the layer input is assigned to different work nodes (mobile devices) to achieve the minimum total execution time.
They used a VGG-16 model, pretrained on ImageNet.
MoDNN was implemented on the LG Nexus 5  (2 GB memory, 2.28 GHz processor, Android 4.4.2).
They reported having successfully accelerated DNN computations by 2.17-4.28 times with 2 to 4 mobile devices.

Stahl et al. perform distributed execution of all layers of a DNN while jointly minimizing
computation, memory usage, and communication overhead \cite{Stahl19DDNN}.
They report even memory distribution across 6 edge-devices with 100 Mbit connections running YOLOv2.
They propose a communication-aware layer fusion scheme that reduces communication cost by 14.8\% and speeds up inference
by 1.15 times.


Distributed computations across edge-devices can lead to undesirable latency.
Hadidi et al. perform partitioning of DNN computations across devices in a robust way using
coded distributed computing (CDC) \cite{10.1145/3316781.3322474}.
They also demonstrate low-latency recovery from errors due to delays in distributed computations.
In other work, Hadidi et al. present a method for dynamic distribution of DNN computations across available devices in order to
achieve real-time performance \cite{8411096}.

Huai et al. have developed a method in which DNN inference required by a robot leverages idle robots in its proximity
by partitioning the model across the robots using latency prediction and optimal partition selection \cite{9060297}.
They report an average speedup of 6 times compared to local execution on the robot or remote execution on the cloud.
 
\section{MACHINE LEARNING APPLICATIONS AT THE EDGE} \label{sec:EdgeMLApplications}
The previous two sections covered techniques that have been developed for training and inference
of ML models on the network edge. 
This section focuses on the actual \emph{applications} of edge-based ML methods
for deploying intelligent systems. 
{\autoref{table:summaryISystems}} summarizes existing research for the development of intelligent systems using deep learning and edge computing.

\subsection{Computer Vision Applications}

\subsubsection{Real-time Video Analytics}
Real-time video analytics systems are an integral part of a wide range of applications, e.g. self-driving cars,
traffic safety and planning, surveillance,  and augmented reality \cite{8057318}.
Until recently, video analytics systems using ML algorithms could only process
about 3 frames per second (FPS) whereas most real-time video cameras stream data at 30 FPS \cite{Kang19realtimevideo}.
Edge computing with IoT cameras has been used to address this problem and provide improved real-time video analytics services.
 
Ananthanarayanan et al. developed a video analytics system called Rocket \cite{8057318}
that produces high-accuracy outputs with low resource costs.
Rocket collects video from different cameras and uses vision processing modules for decoding it.
Each module uses predefined interfaces and application-level optimizations to process video data. 
A resource manager is used to execute data processing tasks on different resource-constrained edge-devices and cloud servers.
A traffic analytics system based on the Rocket software stack has been deployed in Bellevue, WA
to track cars, pedestrians, and bikes.
After processing the data in real-time, it raises an alert if anomalous traffic patterns are detected.
Rocket has been shown to be effective in a variety of applications \cite{Ananthanarayanan:2019:VAK:3307334.3328589},
which include 
\begin{enumerate*}[label=\itshape\alph*\upshape)]
\item  a smart crosswalk for pedestrians in a wheelchair,
 \item  a connected kitchen to pre-make certain food to reduce customer wait-times,
\item traffic dashboard for raising an alarm in abnormal traffic volumes, and
\item retail intelligence for product placement.
\end{enumerate*}
Wang et al. introduced a bandwidth-efficient video analytics architecture based on edge computing
that enables real-time video analytics on small autonomous drones \cite{Wang2018VideoAnalytics}.
DNNs on the drones select interesting frames to be sent to edge-devices.
Periodically, the edge-devices train an SVM to improve the results on the drone DNNs.
These SVMs are transmitted back to the drones where they are used to predict if the DNN output is correct.
In this way, their models are being continuously learned.
Ali et al. designed an edge-based video analytics system using deep learning  to recognize an object in a large-scale IoT video stream \cite{Ali18VedioAnaly}.
There are four stages in this video analytics system:
\begin{enumerate*}[label=\roman*\upshape)]
\item frame loading/decoding and motion detection,
\item  preprocessing,
\item object detection and decomposition, and
\item object recognition. 
\end{enumerate*}
The first three actions are performed on the edge infrastructure and the fourth one in the cloud.
To improve accuracy, this model uses a filter that finds important frames from the video stream
and forwards them to the cloud for recognizing objects.
Their edge-cloud based model was 71\% more efficient than the cloud-based model
in terms of throughput on an object recognition task.

A CNN-based video analytics system  to count vehicles on a road and estimate traffic conditions
without the help of surveillance cameras was designed by Kar et al. \cite{Kar17TrafficVA}.
They considered a vehicle as an edge-device and deployed their model on a dashboard camera on-board the vehicle
to detect other vehicles on the road.
They used 8000 car images to train the model
and then deployed the trained model to identify a moving vehicle and achieved an accuracy of 90\%.

\subsubsection{Image Recognition}
Image recognition refers to the process of extracting meaningful information from a given image, e.g. to identify objects in that image.
Deep learning techniques such as CNNs can be used to detect people, places, handwriting, etc. in an image. 
The prevalence of IoT cameras and mobile devices has increased the importance of improved image recognition techniques
as they are increasingly being used in areas such as wildlife monitoring \cite{7946882}.

Until recently, data would almost always be transferred to the cloud where the images captured by IOT or mobile phones would be processed.
Researchers have increasingly begun to use edge computing techniques to process images close to where they are captured.
For example, there are currently more than 2.5 billion social media users in the world
and millions of photographs and videos are posted daily on social media \cite{DBLP:journals/corr/SouzaCZYCQA15}. 
Mobile phones and other devices can capture high-resolution video which, uploading which may require high bandwidth. 
By processing this data on the edge, the photos and videos are adjusted to a suitable resolution before being uploaded to the Internet \cite{Shi2016EdgeCV}.
Caffe2Go\footnote{\url{https://code.fb.com/android/delivering-real-time-ai-in-the-palm-of-your-hand/}} is a lightweight 
 framework that allows deploying DL systems on a mobile device and helps to reduce the size of the input layer of a DL model.
Personally identifiable information in videos and images can be removed at the edge before they are  uploaded to an external server,
thereby enhancing user privacy.
Liu et al. propose a real-time food recognition system \cite{Liu2018foodrec} that assesses dietary intake by employing deep learning
within the edge computing paradigm. By pre-processing images on edge-devices and classifying them using a CNN on the cloud,
their system achieved highest accuracy, fastest results and minimum energy consumption among all techniques compared.
The main function of the edge-device is to identify a blurry image taken by the user. 
After processing the food image on the edge-device, a clear image is sent to the cloud server for further processing.

Precog is a novel technique designed by Drolia et al. for prefetching and caching to reduce image recognition latency for mobile applications \cite{Drolia17ImageReco}.
On a dataset of mobility traces of museum visitors augmented by paining recognition requests,
Precog was shown to
\begin{enumerate*}[label=\itshape\alph*\upshape)]
\item reduce latency by 5 times,
\item reduce edge-server usage by 5 times, and
\item increase recall by $15\%$.
\end{enumerate*}
Precog employs a method of predicting the next classification request and caching relevant parts of a trained classifier on end-devices. This reduces offloads to the cloud. edge-servers estimate probabilities for the next request from connected end-devices. This estimation is done via a Markov model based on the requests that the edge-server has already received from the devices. Based on these predictions, Precog prefetches parts of trained classifiers and caches them on the end-devices. These pre-fetched parts serve as smaller models that speed-up inference on the end-devices and reduce network communication as well as cloud processing. 

Tuli et al. introduced a deep learning-based real-time object detection system using IoT, fog, and cloud computing \cite{Tuli19ObjectDetection}.
They used the YOLOv3 architecture \cite{yolov3} by training on the COCO dataset to evaluate their system.
The authors have developed an open-source fog-cloud deployment system called EdgeLens\footnote{\url{https://github.com/Cloudslab/EdgeLens}} and demonstrated the capability of this system by deploying object detection YOLO software on multiple Raspberry Pi devices and cloud VMs.

\subsection{Automatic Speech Recognition}
There is immense community interest in developing an offline speech recognition system that supports a digital voice-assistant without the help of the cloud.
Limited-vocabulary speech recognition, also known as keyword spotting is one method to achieve
offline speech recognition \cite{Warden2018LimitedVocabulary}.    
A typical keyword spotting system has two components: 
\begin{enumerate*}[label=\itshape\alph*\upshape)]
\item a feature extractor to gather the necessary features from the human voice and
\item a neural network-based classifier which takes voice features as input and generates a probability for each keyword as output.
\end{enumerate*}
DNN-based keyword spotting systems are not easily deployable on resource-constrained devices.
EdgeSpeechNets is a highly efficient DNN for deploying DL models on  mobile phones
or other consumer devices for human voice recognition \cite{lin2018edgespeechnets}.
It achieved higher  accuracy (approx. 97\%) than state-of-the-art DNNs with a memory footprint of about
1MB on the Google Speech Commands dataset.
When deployed on the Motorola Moto E phone with a 1.4 GHz Cortex-A53 mobile processor as the edge-device,
EdgeSpeechNets used 36 times fewer mathematical operations resulting in 10 times lower prediction latency,
a 7.8 times smaller network size, and a 16 times smaller memory footprint than state-of-the-art DNNs. 
Chen et al. introduced a small-footprint keyword spotting technique based on DNNs called Deep KWS,
which is suitable for mobile edge-devices \cite{Chen2014kWSpotting}. 
Deep KWS has three components: a feature extractor, a deep neural network, and a posterior handling module.
A feature vector extracted from audio input is fed to the DNN
to generate frame-level posterior probabilities scores.
Finally, the posterior handling module uses these scores to generate the final output score of every audio frame to recognize the audio. With respect to traditional HMM based solutions, their model achieved a relative improvement of $45\%$.

\subsection{User Privacy}
The widespread use of IoT devices has boosted personal data production.
However, inadequate security protections on these devices has simultaneously increased the potential for misuse of user data.
Osia et al. designed a hybrid architecture that works with edge and cloud servers collaboratively to protect user privacy.
All personal data is collected and processed on personal edge-devices to remove sensitive information \cite{8364651}.
Only data that is free of sensitive information is  sent to the cloud server for further processing.
They developed a privacy preserving, hybrid edge-cloud system that achieved
$93\%$ accuracy on a gender classification task.
Das et al. have developed a distributed privacy infrastructure 
that can automatically discover nearby IoT devices and notify users about
what data is collected about them and how it is used \cite{8014915}.
Users can then configure privacy preferences (e.g. opt-in/out of any service that uses sensitive data)
which can then be deployed on the IoT devices.

\subsection{Fraud Detection}
With the increase in data being generated by IoT and smart devices, incidents of data fraud and theft are also increasing.
Machine learning is  being used to prevent data falsification and to authenticate data validity.
Ghoneim et al. developed a new medical image forgery detection framework that can identify altered
or corrupted medical images \cite{8337892}.
They used a multi-resolution regression filter on a noise map generated from a noisy medical image
and then used SVM and other classifiers to identify corrupted images.
The first part of their algorithm (that creates a noisy map from a medical image) is done on an edge computing device,
and the rest is done on a cloud server.
This distributed approach decreases the time for data processing as well as bandwidth consumption. 

\subsection{Creating New Datasets}
Feng et al. designed an edge-based architecture that produces a labeled training dataset
which can be used to train a ML model \cite{8567663}.
Based on the amount of training data collected at any stage,
they train progressively more complex ML models at the edge to perform early discard of irrelevant data.
This reduces the amount of labeling that a human has to perform by two orders of magnitude
compared to a brute-force approach.

\subsection{Autonomous Vehicles}
An autonomous vehicle on average generates more than 50 GB of data every minute\footnote{\url{https://datafloq.com/read/self-driving-cars-create-2-petabytes-data-annually/172}}. 
 This data must be processed in real-time to generate driving decisions. 
 The bandwidth of an autonomous vehicle is not large enough for transferring this enormous amount of data to remote servers.
Therefore, edge computing is becoming an integral part of autonomous driving systems.
A novel design of a cognitive internet of vehicles has been proposed by
Chen et al.  \cite{chen2018cognitive}, which they discuss from three perspectives,
intravehicle, intervehicle, and beyond-vehicle networks.
Liang et al. use reinforcement learning for managing network resources by
learning the dynamics of vehicular networks \cite{liang2018toward}.
DeepCrash \cite{8863487} is cloud-based deep learning Internet of Vehicles (IoV) system which can detect
and report a collision event so that timely emergency notifications can be generated.
They report a $96\%$ collision detection accuracy and almost 7 seconds emergency notification latency. 
Hochstetler et al. have shown that it is possible to process real-time video and
detect objects using deep learning on a Raspberry Pi combined with an Intel Movidius Neural Compute Stick \cite{Hochstetler18AV}.
They reported that their embedded system can independently process feeds from multiple sensors in an autonomous vehicle.
Navarro et al. designed a pedestrian detection method for autonomous vehicles \cite{Navarro2016AML}. 
A LIDAR sensor gathers data to detect pedestrians and  features are extracted from this data.
These features include stereoscopic information, the movement of the object,
and the appearance of a pedestrian (local features like Histogram of Oriented Gradients). 
Using the features obtained from raw LIDAR data, an $n$-dimensional feature vector is generated to represent an object on the road. 
This feature vector is used as input for the ML model to detect pedestrians.
They use a Nuvo-1300S/DIO computer to run the ML model inside an autonomous vehicle and
report $96.8\%$ accuracy in identifying pedestrians.
STTR is a smart surveillance system for tracking vehicles in real-time \cite{Xu2018STTR}.
It processes camera streams at the edge and stores space-time trajectories of the vehicles (instead of raw video data)
which reduces the size of the stored data.
Vehicle information can be found by querying these space-time trajectories,
which can be used to help identify suspicious vehicles after traffic accidents.
Tesla has deployed auto-pilot and smart-summon technologies on its autonomous vehicles \cite{karpathy2019pytorchtesla}.
Since vehicles need to detect multiple kinds of objects on the road, they use shared backbone deep networks to reduce computation.
Different object detectors are then trained on top of the backbone networks. To enable both training at scale as well as real-time
inference, they use customized hardware that can execute 144 trillion int8 operations per second operating at less than 100 W.

\subsection{Smart Homes and Cities}
Homes equipped with intelligent systems built using numerous resource-constrained devices 
are increasingly being designed \cite{Harper03SM}.
An important goal in the design of such smart homes is ensuring the safety of children and the elderly.
A fall detection system that generates an alert message when an object falls has been developed by Hsu et al. \cite{7988590}.
Their approach has three steps,
\begin{enumerate*}[label=\itshape\alph*\upshape)]
\item a skeleton extraction performed followed by ML prediction model to detect falls, 
\item video and image compression, for which a a Raspberry Pi is used, and
\item fall detection using ML on the cloud, following which users are notified in appropriate cases.
\end{enumerate*}
Tang et al. designed a hierarchical distributed computing architecture for a smart city
to analyze big data at the edge of the network,
where millions of sensors are connected \cite{7874167}. 
They used SVMs in a smart pipeline monitoring system to 
detect hazardous events to avoid potential damages.
For example, if a house in the smart city experiences a leakage or a fire in the gas delivery system,
edge-devices will detect the threat and quickly shut down the gas
supply of that without the need for any centralized data processing.
An edge-based energy management framework for smart homes 
that improves the use of renewable energy to meet the requirements of IoT applications
has been developed by Chang et al. \cite{8644647}.
Since sunlight is the main source of renewable energy,
they used a weather prediction model to predict weather impacting solar energy generation.
After obtaining forecast information,
an energy scheduling module generates a cost-effective energy utilization schedule for the smart home. 
A Raspberry Pi was used to run their framework at the location of its users,
helping to protect privacy-sensitive data.    	
Park et al. developed an edge-based fault detection system for smart factory settings \cite{s18072110}.
An LSTM recurrent neural network was deployed on an edge-device
(with 1 GB memory and 16 GB flash storage) to detect faults in the working sequence of a robotic arm.


\subsection{Edge AI for Human Safety}
To improve pedestrian security, Miraftabzadeh et al. introduced a ANN-based technique
that uses deep network embeddings as feature vectors for person re-identification in crowded scenarios \cite{8063888}.
edge-devices first try to match the embedding of a detected face with locally stored embeddings.
If no match is found, the embedding is sent for matching and storage in the cloud.

Liu at al. propose an edge-based system that combines audio, video and driving behavior data to detect attacks and potentially harmful events in ride-sharing services \cite{Liu2018Saferide}.
Driver and passenger smartphones act as edge-devices that can trigger video data to be sent to the cloud
for subsequent analysis by a trained CNN model.
The video data can be compressed at the edge to save cloud upload bandwidth.

A framework for an intelligent surveillance system using IoT, cloud computing,
and edge computing has been designed by Dautov et al. \cite{Dautov2018MetropolitanIS}.
Their prototype Metropolitan Intelligent Surveillance System (MISS),
which has been tested on a single-camera testbed, a cloud testbed, and in an edge-cluster setup,
processes sensitive data at the edge of the network to enhance data security and
reduces the amount of data transferred through the network.

\subsection{Augmented Cognition}
Researchers are now exploring how deep learning and edge computing
can be used for augmenting human cognition to create adaptive human-machine collaboration
by quickly giving expert guidance to the human for unfamiliar tasks and for amplifying the human's memory
capabilities \cite{Satya19ACognition}.
Such techniques promise to transform the  way humans with low cognitive abilities
can perform both routine and complex tasks,
but questions pertaining to security, privacy and ethics need to be addressed before such systems are deployed.

\begin{table}[t]
  \centering
  \scriptsize
  \caption{Summary of major edge machine learning applications.}
  \label{table:summaryISystems}
  \renewcommand{\arraystretch}{1.4}
  \begin{tabular}{|p{2cm}|c|p{7cm}|} 
    \hline
        {Application} & {System} & {Short Summary}\\
        \hline
         {~} &  {Rocket \cite{8057318}} &  Real-time video analytics for traffic control, surveillance, and security\\\cline{2-3}
           {~} & JITL \cite{Wang2018VideoAnalytics} & Real-tie video analytics on small autonomous drones\\\cline{2-3}
           {~} &  Edge Enhanced Deep Learning \cite{Ali18VedioAnaly} & Recognize an object in
a large-scale IoT video stream \\\cline{2-3}
{\makecell{Video \\ Analytics}} &  Traffic Estimation \cite{Kar17TrafficVA} & Real-time Traffic Estimation using vehicle as Edge Nodes \\\cline{2-3}
         
      {} & Dietary Assessment System \cite{Liu2018foodrec}  & Real-time food recognition system that help dietary \\\cline{2-3}
      {} & Precog \cite{Drolia17ImageReco} & Reduce image
recognition latency for mobile applications \\\cline{2-3}
{} & EdgeLens \cite{Tuli19ObjectDetection} & Real-time object detection \\\cline{2-3}
	\hline
	{\makecell{Speech \\Recognition}} & EdgeSpeechNets \cite{lin2018edgespeechnets} & Recognition human voice \\\cline{2-3}
	{} & Deep KWS \cite{Chen2014kWSpotting} & Small-footprint keyword spotting technique \\\cline{2-3}
	\hline
	{~} & Hybrid edge-cloud privacy system \cite{8364651} & Protect user privacy when upload data to the cloud\\\cline{2-3}
		{\makecell{Privacy}} & Notify data collection \cite{8014915}& Notify users about what data is collected from the IoT device\\\cline{2-3}
	\hline
	{\makecell{Fraud\\ Detection}} & Forgery detection \cite{8337892}& Authenticate medical image data validity\\\cline{2-3}
	\hline
	{\makecell{Creating New \\ Datasets}} & Eureka \cite{8567663} & Labeling training datasets\\\cline{2-3}
	\hline
	{~} & Cognitive internet \cite{chen2018cognitive} & Used to build vehicle networks \\\cline{2-3}
	{~} & DeepCrash \cite{8863487} & Internet of Vehicles (IoV) system\\\cline{2-3}
	{\makecell{Autonomous \\Vehicles}} & Pedestrian detection \cite{Navarro2016AML} & Pedestrian detection method for autonomous vehicles \\\cline{2-3}
	{~} & STTR \cite{Xu2018STTR} & Smart surveillance system for tracking vehicles \\\cline{2-3}
		{~} & Tesla auto-pilot \cite{karpathy2019pytorchtesla} & Detect multiple kinds of objects on the road \\\cline{2-3}
	\hline
	{~} & Fall detection \cite{7988590} & A system that generates an alert message when an object falls \\\cline{2-3}
	{~} & City Data analytic \cite{7874167} & Detect the threat by analyzing big data  \\\cline{2-3}
	{\makecell{Smart \\ Homes/Cities}} & Energy management \cite{8644647} & A framework for smart homes that improves the use of renewable energy\\\cline{2-3}
	{~} & Fault detection \cite{s18072110} & Detect faults in a smart factory \\\cline{2-3}
	\hline
	{~} & Person identification \cite{8063888} & Identify a person in crowded scenarios\\\cline{2-3}
	{\makecell{Human \\Safety}} & Threat detection  \cite{Liu2018Saferide} & detect attacks and potentially harmful events in ride-sharing services\\\cline{2-3}
	{~} & MISS \cite{Dautov2018MetropolitanIS} & Processes sensitive data at the network edge and enhance seurity \\\cline{2-3}
	\hline
		{\makecell{Augmented \\Cognition}} & Augmenting cognition \cite{Satya19ACognition} & Augmenting human cognition to create adaptive human-machine collaboration \\\cline{2-3}
	\hline
  \end{tabular}
\end{table}

\section{MACHINE LEARNING FRAMEWORKS, SOFTWARE, AND HARDWARE}\label{sec:framewswhw}
The adoption of machine learning models as de facto solutions in an increasingly expanding list of applications has been accompanied by
the development and evolution of many ML frameworks.
In the resource constrained end-edge-cloud architecture,
these developments have been mirrored by customized ML frameworks
which facilitate the building of lightweight  models as well as distributed learning and inference.
We first discuss (\autoref{subsec:frmsftw}) the most widely used frameworks and the software ecosystem used to build and deploy
ML models on the device-edge-cloud architecture and provide a summary in \autoref{table:framework}.
The hardware used in intelligent edge applications is described after that (\autoref{subsec:hardware})
and summarized in  \autoref{table:HardWare}.

\subsection{Frameworks and Software}\label{subsec:frmsftw}

{\em TensorFlow} is a popular machine learning framework, developed by Google.
  TensorFlow Lite\footnote{\url{https://www.tensorflow.org/lite}} is a lightweight implementation of TensorFlow
  for edge-devices  and embedded systems.  
  It has been used for classification and regression on mobile devices. It supports DL without the help of a cloud server
  and has some neural network APIs to support hardware acceleration\footnote{\url{https://www.tensorflow.org/lite/performance/gpu_advanced}}.
  It can be run on multiple CPUs and GPUs and is therefore well-suited for distributed ML algorithms.
  The main programming languages for this framework are Java, Swift, Objective-C, C, and Python.
  A performance evaluation study of TensorFlow Lite by Zhang et al. showed that it occupied only
  84MB memory and took 0.26 seconds to execute an inference task using MobileNets on a Nexus 6p mobile device \cite{Zhang18pCAMP}.

  {\em Caffe2}\footnote{Caffe2 is now a part of \href{https://pytorch.org/}{Pytorch}.}
  is a fast and flexible deep learning framework developed by Facebook. Caffe2Go is a lightweight and modular framework built on top of Caffe2. Both frameworks provide a straightforward way to implement deep learning models on mobile devices
  and can be used to analyze images in real time\footnote{\url{https://code.fb.com/android/delivering-real-time-ai-in-the-palm-of-your-hand/}}.
  Caffe2Go\footnote{\url{https://caffe2.ai/blog/}} can run on the Android and iOS platforms with the same code. It implements debugging tools by abstracting the neural network math. It uses fewer convolutional layers than traditional neural networks and optimizes the width of each layer to reduce model size. 
  
{\em Apache MXNet}\footnote{\url{https://mxnet.apache.org/}} is a lean, scalable, open-source framework
  for training deep neural networks and deploying them on resource-constrained edge-devices.
    MXNet supports distributed ecosystems and public cloud interaction to accelerate DNN training and deployment.
  It comes with tools which help to tracking, debugging, saving checkpoints, and modifying hyperparameters of DNN models.

{\em CoreML}\footnote{\mbox{\url{https://developer.apple.com/machine-learning/core-ml/}}}
    is an iOS-based framework developed by Apple for building ML models and integrating them with Apple mobile applications.
  It allows an application developer to create an ML model to perform regression and image classification. This framework allows ML to run on edge-devices without a dedicated server. A trained DNN model is translated into CoreML format and this translated model can be deployed using CoreML APIs to make an image classifier inside a mobile phone.
  
{\em ML Kit}\footnote{\url{https://firebase.google.com/products/ml-kit}} is a mobile SDK framework introduced by Google. It uses Google's cloud vision APIs, mobile vision APIs, and TensorFlow Lite to perform tasks like text recognition, image labeling, and smart reply.
  Curukogluall et al. tested ML Kit APIs for image recognition, bar-code scanning, and text recognition
  on an Android device and reported that these APIs recognize different types of test objects such as tea cup, water glass, remote controller,
  and computer mouse successfully \cite{mlkit8554039}.

Introduced by Xnor\footnote{\label{fn:xnorai}\url{https://www.xnor.ai/}},
  {\em AI2GO} helps tune deep leaning models for popular use cases on resource-constrained devices.
  More than 100 custom ML models have been built 
  with this framework for on-device AI inferencing. These custom models have the ability to detect objects, classify foods and many other AI applications \cite{Allan19AI2GoBanchmark}.
  This platform targets specialized hardware which includes the Raspberry Pi,
  Ambarella S5L, Linux and macOS based laptops, and Toradex Apalis iMX6.
   Allan conducted tests to benchmark the AI2GO platform on a Raspberry Pi
  and reported this platform to be 2 times faster
  than TensorFlow Lite in ML inferencing using a MobileNets v1 SSD 0.75 depth model \cite{Allan19AI2GoBanchmark}.

  \begin{table*}[t]
    \scriptsize
  \caption{Machine learning frameworks that have been used on edge-devices.}
  \label{table:framework}
  \begin{center}
  \begin{tabular}{|c|c|c|c|c|}
    \hline
    Framework & \makecell{Core development\\ language} & Interface & \makecell{Part running \\ on the edge} & \makecell{Applications}\\
    \hline
    \makecell{TensorFlow Lite\\ (Google)} & \makecell{C++, Java} & \makecell{Android, iOS\\Linux} & \makecell{TensorFlow Lite\\NN API} & \makecell{computer vision \cite{Wang2018VideoAnalytics}, \\speech recognition \cite{TFLitteExamples, AdafruitTFLite}} \\
    \hline
    \makecell{Caffe2, Caffe2Go\\ (Facebook)} & C++ & \makecell{Android \\
iOS} & NNPack & \makecell{image analysis, \\ video analysis \cite{Kangcaffee2inproceedings}} \\
    \hline
       	Apache MXNet & \makecell{C++, R\\Python} & \makecell{Linux\\ MacOS\\ Windows} & Full Model &\makecell{object detection\\ recognition \cite{Nikouei18Surveillance}} \\ 
    \hline
  	\makecell{ Core ML2\\ (Apple)} & Python  & iOS & CoreML & \makecell{image analysis \cite{CML2019object}} \\ 
    \hline
    \makecell{ML Kit\\(Google)} & \makecell{C++\\ Java} &\makecell{Android \\
iOS} &Full Model  &\makecell{image recognition,\\ text recognition, \\bar-code scanning \cite{mlkit8554039}}  \\
    \hline
    \makecell{AI2GO} & \makecell{C, Python\\ Java, Swift} &\makecell{Linux \\
macOs} & Full Model &\makecell{object detection\\ classification \cite{Allan19AI2GoBanchmark}}  \\
    \hline
    DeepThings & C/C++ & Linux & Full Model &\makecell{object detection \cite{Zhaoarticle}}  \\
    \hline
    DeepIoT & Python & Ubilinux &Full Model &\makecell{human activity\\
recognition\\ user identification \cite{Yao2017journals} }  \\
    \hline
    DeepCham & \makecell{C++ \\Java} &\makecell{Linux\\ Android} &Full Model &\makecell{object recognition \cite{Li7774674}}  \\
    \hline
    SparseSep & - &\makecell{Linux \\ Android} &Full Model &\makecell{mobile object recognition\\ audio classification \cite{bhattacharya2016spars}}\\
    \hline
    DeepX & \makecell{C++, Java,\\ Lua} &\makecell{Linux \\ Android} &Full Model &\makecell{mobile object recognition\\ audio classification \cite{Lane16Deepx}}\\
    \hline
    Edgent & - &\makecell{Ubuntu} &\makecell{Major part \\ of the DNN} &\makecell{image recognition \cite{Li2018EdgeOndevice}}\\
\hline
    daBNN & \makecell{Arm Assembly, \\C++, Java} &\makecell{Mobile OSs \\( Android and others)} &Full model &\makecell{Computer vision \cite{Zhang2019daBNNAS}}\\
    \hline
    \makecell{TensorFlow Federated\\(Google)} & Python & iOS, Android & \makecell{Distributed\\ training} & computer vision \cite{Lin17DSGD} \\
    \hline
    \makecell{EdgeML Library\\(Microsoft Research)} & C++ & - & - & image recognition \cite{pmlr-v70-gupta17a} \\
    \hline
    CONDENSA & Python & \makecell{Linux, Windows,\\ MacOS} & Full Model & \makecell{language modeling \\ and computer vision \cite{joseph19CONDENSA}} \\
    \hline
  \end{tabular}
  \end{center}
\end{table*}

{\em DeepThings} is a framework for adapting CNN-based inference applications on
  resource-constrained devices \cite{Zhaoarticle}.
  It provides a low memory footprint of convolutional layers by using Fused Tile Partitioning (FTP). 
FTP divides the CNN model into multiple parts and generates partitioning parameters.
These partitioning parameters with model weights are then distributed to edge-devices. 
When all edge-devices complete their computational tasks, a gateway device collects the processed data and generates results.
The authors deployed YOLOv2 using DeepThings on Raspberry Pi 3 devices to demonstrated the deployment capability of this framework on IoT devices.
  
{\em DeepIoT} is a framework that shrinks a neural network into smaller dense matrices
  but keeps the performance of the algorithm almost the same \cite{Yao2017journals}.
This framework finds the minimum number of filters and dimensions required by each layer and reduces the redundancy of that layer.
Developed by Yao et al., DeepIoT can compress a deep neural network by more than 90\%,
shorten execution time by more than 71\%, and decrease energy consumption by 72.2\% to 95.7\%. 

Li et al. introduced a framework called {\em DeepCham} which allows developers to deploy DL models on mobile environments with the help of edge computing devices \cite{Li7774674}.
  DeepCham is developed for recognizing objects captured by mobile cameras, specifically targeting Android devices.

{\em SparseSep}, developed by Bhattacharya and Lane, is a framework for optimizing large-scale DL models
  for resource-constrained devices such as wearable hardware \cite{bhattacharya2016spars}. 
It run large scale DNNs and CNNs on devices that have ARM Cortex processors with very little impact on inference accuracy. 
It can also run on the NVidia Tegra K1 and the Qualcomm Snapdragon processors  
and was reported to run inference 13.3 times faster with 11.3 times less memory than conventional neural networks. 

Lane et al. designed a software accelerator for low-power deep learning inference called \emph{DeepX},
  which allows  developers to easily deploy  DL models on mobile and wearable devices \cite{Lane16Deepx}.
  DeepX dramatically lowers resource overhead by decomposing a large deep model network into unit-blocks.
  These unit-blocks are generated using two resource control algorithms, namely Runtime Layer Compression
  and Deep Architecture Decomposition, and executed by heterogeneous processors (e.g. GPUs, LPUs)
  of mobile phones.    
  
Li et al. developed \emph{Edgent}, 
  a framework for deploying deep neural networks on small devices \cite{Li2018EdgeOndevice}.
  This framework adaptively partitions DNN computations between a small mobile device (e.g. Raspberry Pi)
  and the edge computing devices (e.g. laptops), and
  uses early-exit at an intermediate DNN layer to accelerate DNN inference.
  
{\em  daBNN} is an open-source fast inference framework developed by  Zhang et al., which can implement Binary Neural Networks on ARM devices \cite{Zhang2019daBNNAS}. An upgraded bit-packing scheme and binary direct convolution have been used in this framework to shrink the cost of convolution and speed up inference. This framework is written in C++ and ARM assembly and has Java support for the Android package. This fast framework can be 6 times faster than BMXNet\footnote{\url{https://github.com/hpi-xnor/BMXNet}} on Bi-Real Net 18\footnote{\url{https://github.com/liuzechun/Bi-Real-net}}.   

Joseph et al. developed a programmable system called {\em CONDENSA} which allows developers to design strategies for compressing DL models, resulting significant reduction in both the memory footprint and execution time \cite{joseph19CONDENSA}.  It uses a Bayesian optimization-based method to find compression hyperparameters automatically by exploiting an L-C optimizer to recover the accuracy loss during compression. They reported a 2.22 times runtime improvement over an uncompressed VGG-19 on the CIFAR-10 dataset with up to 65 times memory reduction.

In addition to frameworks covered earlier, the edge ML ecosystem
now also includes operating systems, domain specific programming languages, and
software to facilitate the development of cloud-edge services. 

{\em SeeDot} is a programming language, developed by Microsoft Research, to express ML inference algorithms and to control them at a mathematical-level \cite{Gopinath:2019:CKM:3314221.3314597}.
  Typically, most learning models are expressed in floating-point arithmetic, which are often expensive.
  Most resource-constrained devices do not support floating-point operations. To overcome this, SeeDot generates fixed-point code with only integer operations,
  which can be executed with just a few kilobytes of RAM. This helps SeeDot run a CNN on a resource-constrained micro-controller with no floating-point support.
  SeeDot-generated code for ML classification that uses Bonsai and ProtoNN (see~\autoref{sec:MLAlgoForEdge})
  is 2.4 to 11.9 times faster than floating-point microcontroller-based code.  
  Also, SeeDot-generated code was 5.2-9.8 times faster than code generated by high-level synthesis tools
  on an FPGA-based implementation.  

\emph{AWS IoT Greengrass}\footnote{\url{https://aws.amazon.com/greengrass/}}
  is software which helps an edge-device to run serverless AWS Lambda functions.
  It can be used to run ML inference on an edge-device, filter device data,
  sync data and only transmit important information back to the cloud.
The \emph{Azure IoT Edge}\footnote{\url{https://azure.microsoft.com/en-gb/services/iot-edge/}}
  is a platform that can be used to offload a large amount of work from the cloud to the edge-devices. 
  It has been used to deploy ML models on edge-devices and cloud servers.
  Such workload migration reduces data communication latency and operates reliably even in offline periods.

{\em Zephyr}\footnote{\url{https://www.zephyrproject.org/what-is-zephyr/}} 
  is a real-time operating system with a small-footprint kernel, specially designed for resource-constrained devices.
  This OS supports multiple architectures, such as Intel x86, ARM Cortex-M, RISC-V 32, NIOS II, ARC,
  and Tensilica Xtensa.
  Zephyr is a collaborative project, which is hosted by the Linux Foundation under the Apache 2.0 license. 

  \mbox{\em TVM} is a compiler that has the capability of optimizing code by searching and detecting optimized tensor operators \mbox{\cite{222575}}. This compiler provides end-to-end compilation and optimization stacks which allow the deployment of deep learning on mobile GPUs, FPGA-based devices. This open-sourced compiler was evaluated on different edge devices such as embedded GPU (ARM Mali-T860MP4 GPU), FPGA-based accelerator, and reported a speedup of $1.2$ to $3.8$ times
    over existing hand-optimized libraries based frameworks such as Tensorflow Lite.

\subsection{Hardware}\label{subsec:hardware} 
This section describes the low-power hardware that have been used in edge machine learning systems.
High performance from a deep learning application is only achievable when a ML model is trained with a huge amount of data, often on the order of terabytes.
Computationally rich GPUs, and central CPUs only have the ability to handle such a large amount of data in a reasonable period of time.
This makes deep learning applications mostly GPU-centric. 
However, efforts have been taken to make resource-constrained devices compatible with deep learning and it has been noticed that different types of small devices are being
used to deploy ML at the network edge, including ASICs, FPGAs, RISC-V, and embedded devices. 
\autoref{table:HardWare} lists  devices commonly used to deploy ML systems at the
edge.
 

\begin{table*}[!t]
  \scriptsize
  \caption{Computing devices that have been used for Machine learning at the edge.}
  \label{table:HardWare}
  \renewcommand{\arraystretch}{2}
  \begin{center} 
\scalebox{1}{ 
  \begin{tabular}{|c|c|c|c|c|c|c|c|}
    \hline
    Device type & Name & GPU & CPU & RAM & Flash memory & Power & \makecell{Applications}\\
    \hline
   {~} &  \makecell{Google Coral\\Dev Board} & \makecell{GC7000 Lite\\Graphics +\\Edge TPU\\coprocessor} & \makecell{Quad\\Cortex-A53,\\Cortex-M4F} & \makecell{1 GB\\ LPDDR4} & \makecell{8 GB\\ LPDDR4} & 5V DC & \makecell{image processing \cite{Cass19Coral}} \\
    \cline{2-8}    
    {~} &  \makecell{SparkFun\\ Edge} & - & \makecell{32-bit ARM\\Cortex-M4F\\48MHz} & 384KB & 1MB & 6uA/MHz & \makecell{speech recognition \cite{sparkfun}} \\
    \cline{2-8} 
    {\makecell{(ASICs) (\autoref{subsubsec:ASICsDevices})}} &  \makecell{Intel\\Movidius\\NCS} & \makecell{High\\Performance\\VPU} & \makecell{Myriad 2\\VPU} &1 GB & 4 GB & \makecell{2 trillion\\16-bit ops/s\\within\\500 mW} &\makecell{classification \cite{Marantos8376630} \\ computer vision \cite{Barry7024073, Hochstetler18AV}}\\ 
    \cline{2-8}
   {~} &  \makecell{BeagleBone \\ AI} & - & \makecell{Cortex-A15 \\Sitara AM5729\\ SoC with 4 EVEs} & \makecell{1 GB} & 16 GB & \makecell{-} &\makecell{computer vision \cite{beagleboneai}}\\ 
    \cline{2-8}
   {~} &  \makecell{Eta Compute \\ EMC3531} & - & \makecell{ARM Cortex-M3 \\NXP Coolflux \\DSP} & \makecell{-} & - & \makecell{-} &\makecell{audio, video \\ analysis\textsuperscript{\ref{hw:etadevice}}}\\ 
    \hline      
    {\makecell{(FPGAs) (\autoref{subsubsec:FPGAsDevices})}} &  \makecell{ARM ML} & - & \makecell{ARM ML\\processor} & 1 GB & - & \makecell{4 TOPs/W\\(Tera\\Operations)} &\makecell{image, voice \\ recognition \cite{Wang7505926}}\\ 
    \hline   

    {~} &  \makecell{Raspberry\\ Pi} & \makecell{400MHz\\VideoCore\\IV} & \makecell{Quad\\Cortex A53\\@ 1.2GHz} & \makecell{1 GB\\SDRAM} & \makecell{32 GB} & 2.5 Amp & \makecell{video analysis \cite{Nikouei18Surveillance, Xu8422970}} \\
    \cline{2-8}
        {~} &  \makecell{NVIDIA\\Jetson TX1} & \makecell{Nvidia\\Maxwell\\256 CUDA\\cores} & \makecell{Quad ARM\\ A57/2 MB \\ L2} & \makecell{4 GB\\ 64 bit\\LPDDR4\\25.6 GB/s} & \makecell{16 GB eMMC,\\SDIO, SATA} & 10-W & \makecell{video, image\\analysis \cite{Liu2017DemoAR, Lee17jetsonCV}, \\robotics \cite{Tsur8540206}}\\
    \cline{2-8}
    \makecell{Embedded \\ GPUs (\autoref{subsubsec:EmbeddedDevices})} &  \makecell{NVIDIA\\Jetson TX2} & \makecell{Nvidia\\Pascal\\256 CUDA\\cores} & \makecell{HMP Dual\\Denver 2/2\\MB L2 +\\Quad ARM\\A57/2\\MB L2} & \makecell{8 GB\\128 bit\\LPDDR4\\59.7 GB/s} & \makecell{32 GB eMMC,\\SDIO, SATA} & 7.5-W &\makecell{video, image\\analysis \cite{Liu2017DemoAR, Rungsuptaweekoon}, \\ robotics \cite{JetsontxRobotic}}\\ 
    \cline{2-8}
        {~} &  \makecell{NVIDIA\\Jetson NANO} & \makecell{Nvidia\\Maxwell\\128 CUDA\\cores} & \makecell{Quad ARM\\ A57 MPCore} & \makecell{4 GB\\ 64 bit\\LPDDR4\\25.6 GB/s} & \makecell{16 GB eMMC,\\ 5.1 Flash} & 5 - 10-W &\makecell{computer\\vision \cite{8951147, 9070378}, \\ audio analysis \cite{Gao2020EdgeDRNNEL}}\\ 
    \cline{2-8}
    {~} &  \makecell{OpenMV \\Cam} & - & \makecell{ARM 32-bit\\ Cortex-M7 } & \makecell{512KB} & 2 MB & \makecell{200mA \\@ 3.3V} &\makecell{image processing \cite{openMVCam}}\\ 
    \hline
        
        \makecell{Open (ISAs) (\autoref{subsubsec:ISAdevices})} &  \makecell{RISC-V\\ GAP8} & - & \makecell{nona-core\\32-bit RISC-V\\@250 MHz} & \makecell{16 MiB\\ SDRAM} & - & 1 GOPs/mW &\makecell{image, audio\\processing \cite{Flamand18GAP8}}\\ 
    \hline
    \end{tabular}
} 
  \end{center}
\end{table*}

\subsubsection{Application-specific integrated circuits (ASICs)} \label{subsubsec:ASICsDevices}
The {\em Edge Tensor Processing Unit\footnote{\url{https://cloud.google.com/edge-tpu/}}} (TPU) is an ASIC chip designed by Google for accelerated ML inference on edge-devices.
  It is capable of running CNNs such as MobileNets V1/V2, MobileNets SSD V1/V2, and Inception V1-4
  as well as TensorFlow Lite models.
 In the power efficient mode, the Edge TPU can execute state-of-the-art mobile vision models at 100+ fps \footnote{\url{https://coral.withgoogle.com/docs/edgetpu/faq/}}.
    The {\em Coral Dev Board}, also by Google, uses the Edge TPU as a co-processor to run ML applications. 
    It has two parts, a baseboard and a system-on-module (SOM).
    The baseboard has a 40-pin GPIO header to integrate with various sensors or IoT devices and
    the SOM has a Cortex-A53 processor with an additional Cortex-M4 core, 1GB of RAM and 8GB of flash memory
    that helps to run Linux OS on Edge TPU.

    The {\em Coral USB accelerator}\footnote{\url{https://coral.withgoogle.com/products/accelerator/}} is a device 
    that helps to run ML inference on small devices like the Raspberry Pi.
    The accelerator is a co-processor for an existing system, which can connect to any Linux system using a USB-C port.
    Allan deployed ML inference on different edge-devices, including the Coral Dev Board
    and conducted tests to benchmark  the inference performance \cite{Allan19Benchmark, Allan19MeasureML}.
    The Coral Dev Board performed 10 time faster than Movidius NCS, 3.5 time faster than Nvidia Jetson Nano (TF-TRT) and 31 time faster than Raspberry Pi for
    MobileNetV2 SSD model.

The {\em SparkFun Edge} is a real-time audio analysis device,
  which runs ML inference to detect a keyword, for example, "yes" and responds accordingly\footnote{\url{https://learn.sparkfun.com/tutorials/sparkfun-edge-hookup-guide/all}}.  
  Developed by Google, Ambiq, and SparkFun collaboratively, it's used for voice and gesture recognition at the edge\footnote{\url{https://www.sparkfun.com/products/15170}}
  without the help of remote services. It has a 32-bit ARM Cortex-M4F 48MHz processor with 96MHz burst mode, extremely low-power usage, 384KB SRAM, 1MB flash memory and a dedicated BLE 5 Bluetooth processor.
  It also has two built-in microphones, a 3-axis accelerometer, a camera connector,  and other input/output connectors. This device can run for 10 days with a CR2032 coin cell battery.
  Ambiq Apollo3 is a Software Development Kit is available for building AI applications with the SparkFun Edge.

 Intel's {\em Movidius}\footnote{\url{https://www.movidius.com/}}
  is a vision processing unit which can accelerate deep neural network inferences in resource-constrained devices such as intelligent security cameras or drones. This chip can run custom vision, imaging, and deep neural network workloads on edge-devices without a connection to the network or any cloud backend. 
  This chip can be deployed on a robot placed in rescue operations in disaster-affected areas. The rescue robot can make some life-saving decisions without human help.
  It can run real-time deep neural networks by performing 100 gigaflops within a 1W power envelope. Movidius Myriad 2 is the second generation vision processing unit (VPU) and Myriad X VPU is the most advanced VPU from Movidius to provide artificial intelligence solutions from drones and robotics to smart cameras
  and virtual reality\footnote{\url{https://www.movidius.com/myriadx}}.
  Intel also provides a Myriad Development Kit (MDK) which includes all necessary tools and APIs to implement ML on the chip.
  
  The {\em  Movidius Neural Compute Stick} by Intel is a USB like stick which extends
  the same technology of Intel Myriad (SoC) board.
  This plug and play device can be easily attached to edge-devices running by Ubuntu 16.04.3 LTS (64 bit), CentOS* 7.4 (64 bit), Windows 10 (64 bit), Raspbian (target only), including Raspberry Pi, Intel NUC, etc.
  This device has Intel Movidius Myriad X Vision Processing Unit (VPU) processor,
  which supports TensorFlow, Caffe, Apache MXNet, Open Neural Network Exchange (ONNX), PyTorch, and
  PaddlePaddle via an ONNX conversion. 
  
 The  {\em BeagleBone AI}
is a high-end board for developers building machine-learning and computer-vision applications \cite{beagleboneai}.
    This device is powered by an SoC -- TI AM5729 dual core Cortex-A15 processor featuring 4 programmable real-time units,
    a dual core C66x digital-signal-processor, and 4 embedded-vision-engines core supported through the TIDL (Texas Instruments Deep Learning) ML API.
    It can perform image classification, object detection, and semantic segmentation using TIDL.
    
    {\em ECM3531}\footnote{\label{hw:etadevice}\url{https://etacompute.com/products/}} is a high-efficiency ASIC based on the ARM Cortex-M3 and NXP Coolflux DSP processors for machine learning applications. The name of the processor of this ASIC is Tensai, which can run TensorFlow or Caffe framework. This processor offer 30-fold power reduction in a specific CNN-based image classification.   
    
The    {\em SmartEdge Agile}\footnote{\url{https://www.avnet.com/wps/portal/us/solutions/iot/building-blocks/smartedge-agile/}}
device along with the accompanying {\em Brianium} software
      help build artificial intelligence models and deploy them on resource-constrained devices.
      SmartEdge Agile sits uses Brainium's zero-coding platform
      to deploy a trained intelligent model on the edge \cite{Allan19Agile}. 
  

\subsubsection{Field-programmable gate arrays (FPGAs)} \label{subsubsec:FPGAsDevices}
The {\em ARM ML processor} \cite{ArmML}
  allows developers to accelerate the performance of ML algorithms and deploys inference on edge-devices. 
    The ecosystem consists of 
    \begin{enumerate*}[label=\itshape\alph*\upshape)]
      \item ARM NN\footnote{\url{https://mlplatform.org/}}, an inference engine that provides a translation layer
        that bridges the gap between existing Neural Network frameworks and ARM ML processor, and
      \item ARM Compute Library\footnote{\url{https://www.arm.com/why-arm/technologies/compute-library}},
        an open source library containing functions optimized for ARM processors.
    \end{enumerate*}
    The ARM ML  processor can run high-level neural network frameworks like TensorFlow Lite, Caffe, and ONNX.
    The ARM NN SDK has all the necessary tools to run neural networks on edge-devices.
    This processor is designed for mobile phones, AR/VR, robotics, and medical instruments.
    Lai and Suda designed a set of efficient neural network kernels called CMSIS-NN\footnote{\url{https://github.com/ARM-software/CMSIS_5}}    to maximize the performance of a neural network using limited memory and compute resources of an ARM Cortex-M processor \cite{Lai2018EDL}.
    
Microsoft's {\em Brainwave} is an effort to use FPGA technology to solve the challenges of real-time AI and to run deep learning models in the Azure cloud and on the edge in real-time \cite{MSPBrainwave}. 
To meet the computational demands required of deep learning, Brainwave uses Intel Stratix 10 FPGAs as a heart of the system providing 39.5 TFLOPs of effective performance. 
Most popular deep learning models, including ResNet 50, ResNet 152, VGG-16, SSD-VGG, DenseNet-121, and SSD-VGG are supported Brainwave FPGAs on Azure to accomplish image classification and object detection task at the network edge.
Azure can parallelize pre-trained deep neural networks (DNN) across FPGAs to scale out any service.    

\subsubsection{Embedded GPUs} \label{subsubsec:EmbeddedDevices}
The {\em Raspberry Pi}, a single-board computer developed by the Raspberry Pi Foundation,
  is one of the most common devices used for edge computing.
  It has been used to run ML inference without any extra hardware.
  The Raspberry Pi 3 Model B has a Quad Cortex A53 @ 1.2GHz CPU, 400MHz VideoCore IV GPU, 1GB SDRAM.
  Xu et al. used Raspberry Pi 3 as edge-devices to develop a real-time human surveillance system \cite{Nikouei18Surveillance, Xu8422970}. Their system is able to distinguish between human and nonhuman objects in real-time.  
  It has a micro-SD card slot to support flash memory up to 32 GB.  Xnor.ai has developed a new AI platform to run deep learning models efficiently on edge-devices such as embedded CPUs (e.g. Raspberry Pi), phones, IoT devices, and drones without using a GPU or TPU \cite{Laborde19RaspPi, Welsh19TrueAI}.
    
The {\em Nvidia Jetson} is an embedded computing board that can process complex data in real-time.
  The Jetson AGX Xavier can operate with a
  30W power supply and perform like a GPU workstation for edge AI applications. 
{\em Jetson TX1}, {\em Jetson TX2} and {\em Jetson NANO} are embedded AI computing devices powered by the Jetson.
These three small, but powerful,  computers are ideal for implementing
an intelligent system on edge-devices such as smart security cameras, drones, robots, and portable medical devices. 
JetPack is an SDK for building AI applications with the Jetson.
This SDK includes TensorRT, cuDNN, Nvidia DIGITS Workflow, ISP Support, Camera imaging, Video CODEC, Nvidia VisionWorks, OpenCV, Nvidia CUDA,
and CUDA Library tools for supporting ML.
It is also compatible with the Robot Operating System (ROS\footnote{\url{https://www.ros.org}}).

{\em OpenMV Cam}\footnote{\url{https://openmv.io/collections/cams/products/openmv-cam-m7}} is a small, low-powered camera board. This board is built using an ARM Cortex-M7 processor
    to execute machine vision algorithms at 30 FPS.
    This processor can run at 216 MHz and has 512KB of RAM, 2 MB of flash, and 10 I/O pins.
    The main applications of this device are face detection, eye tracking, QR code detection/decoding,
    frame differencing, AprilTag tracking, and line detection \cite{openMVCam}.

\subsubsection{Open Instruction Set Architecture (ISA)} \label{subsubsec:ISAdevices}
The  GAP8  \cite{GAP8GreenWaves} is a microprocessor conforming to the RISC-V
open set architecture \cite{riscvreader}.
With 9 cores capable of running 10 GOPS at the order of tens of mW, 
    this 250 MHz processor is designed to accelerate CNNs for the edge computing and IoT market.
    The TF2GAP8 is a tool that can automatically translate TensorFlow CNN applications to GAP8 source \cite{GAP8TF2GAP8}.

\section{CHALLENGES AND FUTURE DIRECTIONS} \label{sec:ChallengesAndFD}
In order to fully utilize the benefits offered by edge intelligence, a number of issues that
significantly inhibit more widespread adoption of edge-based machine learning applications need to be addressed.\\    

\noindent \textbf{Building new datasets.} An important characteristic of deep learning algorithms
is their ability to train using labeled as well as unlabeled  input data.
The availability of large amounts of unlabeled data generated by edge-servers and end-devices 
provides a faster way of building new datasets.
For low-data scenarios, data augmentation can be employed \cite{NWEKE2018233}.
Data augmentation uses a small amount of data (transferred from sensor to edge-servers or from edge-servers to cloud servers)
to generate new data.
Augmentation helps ML models avoid overfitting issues by generating enough new training data \cite{NWEKE2018233}.
However, the performance of data augmentation by edge-devices or cloud servers needs to be evaluated before its use
in a learning model, especially for small datasets.

The use of various types of IoT sensors  creates heterogeneous environments in edge-based intelligent systems.
To deal with the diversity of data in edge environments,
ML algorithms need to learn using types of data that have different features like image, text, sound, and motion.
Multimodal deep learning is adapted to learn features over multiple modalities such as audio and video \cite{Ngiam2011MultimodelDL} for the high-velocity heterogeneous data generation that is common in intelligent edge settings.
An additional challenge arises when the data keep changing over time and the joint probability between classes and the generated data changes because of seasonality, periodicity effects or hardware/software failure. An important area of research is to create novel machine learning algorithms for handling the non-stationary data streams generated by IoT sensors \cite{7296710}.
\newline

\noindent \textbf{Centralized vs distributed training.} 
To handle the enormous amount of data produced by IoT sensors, researchers have designed
an edge-based distributed learning algorithm (\autoref{sec:distrInference}) over distributed computing hierarchies.
The  architecture consists of the cloud servers, the edge-servers, and end-devices like IoT sensors.
Such  algorithms provide acceptable accuracy with datasets that are naturally distributed, for example, fraud detection and market analysis.
However, the influence of the heterogeneity of data on the accuracy of a distributed model is an open research issue \cite{PeteiroBarral2013}. 

Since training a deep learning model on edge-devices is difficult (due to their limited memory and computational capabilities), most existing ML systems use the cloud for training.
Some attempts have been made  to train models on edge-devices (such as by using model pruning and model quantization)
but pruned edge-trained models often have lower accuracy, and therefore designing power-efficient algorithms for training neural networks on edge-devices is an active research area.
There remains continued interest in developing new methods and frameworks that map sections of a deep learning model onto the distributed computing architecture
and also in exploring the use of specialized hardware (such as ARM ML processors) to speed up deep learning training and inference.

Computation results and data-sharing 
across different edge-devices is key component to establish an effective ML-edge distributed system.
Novel networking paradigms that are `computation-aware' are highly desirable for
building such data-sharing distributed systems.
5G networks, which provide the ultra-reliable low-latency communication (URLLC) services, are a promising area to integrate with edge computing. 5G should help to establish more control over the network resources for supporting on-demand interconnections across different edge-devices. The adaptation of the software-defined network and network function virtualization into 5G networks to control distributed ML settings will be an appealing research area for future ML researchers.\\


\noindent \textbf{Trust and explainability.} A major concern in the ML and wider community is
of the accountability and transparency of ML systems.
The current black-box-like decision-making process of ML-based AI systems 
has raised questions about inherent algorithmic bias \cite{Ribeiro2016WhySI}.
This has led to the development of 
\emph{Explainable AI} (XAI) and {\em interpretable AI}, which seek to 
provide the reasons why a particular decision was reached by the AI system. 
An overview of issues in XAI can be found in the survey by Adadi and Berrada \cite{adadi2018peeking}.

Sensitive domains such as health-care, law-enforcement, and jurisprudence 
require high accuracy as well as explainability.
However, deep learning models deployed on edge-devices
are inherently less accurate given the resource-constrained environment.
Therefore, a primary challenge facing the community is to explore how to ensure that
no permanently harmful decision is  made using an intelligent edge system.

Due to limited sizes and inherent biases in datasets, ML models can treat different subsets
of the population differently and even unfairly \cite{simonite2019best}. 
This has led to the development of fairness-enhanced ML models.
However, this aspect of ML remains under-explored and different methods of introducing fairness
are still affected by dataset bias  \cite{10.1145/3287560.3287589}.
Introducing fairness as a criterion during ML model training raises some important issues.
For example, Biswas and Rajan discuss how  fairness with respect to one attribute can lead to lesser fairness
with respect to another attribute and also that optimizing for fairness can lead to reduction in performance \cite{biswas20machine}.
Another important focus increasingly is incorporating fairness criteria and their optimization in
ML software libraries and frameworks.\\

\noindent {\bf Novel Topics.}
With increase in the dedicated hardware for ML,
an important direction of future work is the development of compilers,
such as Glow  \cite{rotem2018glow}, that optimize neural network graphs for heterogeneous hardware.
Li et al. provide a detailed survey of the architectures of deep learning compilers \cite{li2020deep}.
Data produced by resource-constrained end-devices in a decentralized distributed setting is always vulnerable to security threats. 
Edge-based blockchain technology has great potential to prevent those threats and transfer sensitive data over a decentralized edge infrastructure.
However, deploying blockchains in resource-constrained settings is challenging due to the huge computing power and energy requirements in blockchain mining \cite{wang2020convergence}. 
\newline

\section{CONCLUSION} \label{sec:conclusion}
Edge-based ML is a fast-growing research area with numerous challenges and opportunities.
Using edge-devices for ML has been shown to improve not only the privacy and security of  user data
but also system response times.
This article provides a comprehensive overview of techniques  pertaining to the training and
deployment of ML systems at the network edge.
It highlights several new ML architectures that have been designed specifically for resource-constrained
edge computing devices and reviews important applications of edge-based ML systems.
Widely adopted frameworks essential for developing edge-based deep learning architectures
as well as the resource-constrained devices that have been used to deploy these models are described.
Finally,  challenges in edge-based ML are listed and several directions for future work are outlined.

\bibliographystyle{ACM-Reference-Format}
\bibliography{bibfile}

\end{document}